\documentclass[11pt]{article}

\usepackage[T1]{fontenc}
\usepackage[utf8]{inputenc}
\usepackage{lmodern}
\usepackage{microtype}
\usepackage[letterpaper,margin=1in]{geometry}
\usepackage{amsmath,amsfonts,amssymb,amsthm,bm}
\usepackage{mathtools}
\usepackage{graphicx}
\usepackage{subcaption}
\usepackage{booktabs}
\usepackage{enumitem}
\usepackage{placeins}
\usepackage[dvipsnames]{xcolor}
\usepackage[colorlinks=true,linkcolor=blue!55!black,citecolor=green!35!black,
            urlcolor=blue!55!black]{hyperref}

\numberwithin{equation}{section}
\theoremstyle{plain}
\newtheorem{theorem}{Theorem}[section]
\newtheorem{lemma}[theorem]{Lemma}
\newtheorem{proposition}[theorem]{Proposition}

\theoremstyle{definition}
\newtheorem{definition}[theorem]{Definition}
\newtheorem{assumption}[theorem]{Assumption}
\newtheorem{example}[theorem]{Example}
\theoremstyle{remark}
\newtheorem{remark}[theorem]{Remark}
\newenvironment{namedproof}[1]{\begin{proof}[#1]}{\end{proof}}

\newcommand{\email}[1]{\href{mailto:#1}{\texttt{#1}}}
\newenvironment{keywords}{\par\medskip\noindent\textbf{Keywords.}\ }{\par}
\newenvironment{AMS}{\par\smallskip\noindent\textbf{AMS subject classifications.}\ }{\par}

\title{Kernel-Based Learning of Analytic Operators and Parametric
PDEs\thanks{This material is based upon work supported by the European Commission (Horizon Europe grant
ELIAS 101120237). Co-funded by the European Union (ERC, SAMPDE, 101041040). Views and opinions expressed are however those of the authors only and do not necessarily reflect those of the European Union or the European Research Council. Neither the European Union nor the granting authority can be held responsible for them. The research was supported in part by the MIUR Excellence Department Project awarded to Dipartimento di Matematica, Università di Genova, CUP D33C23001110001.}}

\author{Giovanni S. Alberti\thanks{MaLGa Center, DIMA,
University of Genoa, Via Dodecaneso 35, 16146 Genoa, Italy
(\email{giovanni.alberti@unige.it}, \email{matteo.santacesaria@unige.it}).}
\and Markus Holzleitner\thanks{Institute for Machine Learning, Johannes
Kepler University Linz, Altenberger Stra\ss e 69, 4040 Linz, Austria
(\email{holzleitner@ml.jku.at}).}
\and Lorenzo Rosasco\thanks{MaLGa Center, DIBRIS,
University of Genoa, Via Dodecaneso 35, 16146 Genoa, Italy
(\email{lorenzo.rosasco@unige.it}).}
\and Matteo Santacesaria\footnotemark[2]}

\newcommand{\X}{\mathcal{X}}
\newcommand{\Hs}{\mathcal{H}}
\newcommand{\Y}{\mathcal{Y}}
\newcommand{\I}{\mathbb{I}}

\newcommand{\Real}{\mathbb{R}}
\newcommand{\EXP}{\mathbb{E}}

\newcommand{\norm}[1]{\left\lVert#1\right\rVert}
\newcommand{\innerpro}[1]{\left\langle#1\right\rangle}

\newcommand{\tpsi}{\widetilde{\psi}}

\newcommand{\abs}[1]{\left|#1\right|}

\newcommand{\Fell}{F_{*}^{\mathrm{ell}}}

\makeatletter
\newcommand{\dlmf}[1]{%
\cite[%
  \def\nextitem{\def\nextitem{, }}%
  \@for \el:=#1\do{\nextitem\href{http://dlmf.nist.gov/\el}{(\el)}}%
]{olver2010nist}%
}
\makeatother

\newcommand{\maybeincludegraphics}[2][]{%
  \IfFileExists{#2}{\includegraphics[#1]{#2}}{%
    \fbox{\parbox[c][0.22\textheight][c]{0.8\linewidth}{\centering
      Figure file `\texttt{\detokenize{#2}}' not supplied}}}}

\begin{document}
\maketitle

\begin{abstract}
We study kernel-based learning of analytic operators from finite samples,
with a focus on how the regularity of the target should determine the
choice of the kernel. We establish embeddings of analytic operators into
vector-valued reproducing kernel Hilbert spaces and derive finite-sample,
high-probability guarantees that quantify the uncertainty due to random
sampling: in the noiseless setting, suitable analytic kernels yield rates
exponential in a fractional power of the sample size, while Sobolev-type
kernels yield polynomial rates. We apply the theory to the solution
operator of a parametric elliptic PDE, where the training data are
themselves finite-element approximations, and obtain bounds that separate
the sampling and discretization contributions to the error and prescribe
how the mesh size should be coupled to the sample size. The results are
illustrated numerically.
\end{abstract}

\begin{keywords}
operator learning, vector-valued reproducing kernel Hilbert spaces, kernel
ridge regression, analytic operators, parametric elliptic PDEs, learning rates
\end{keywords}

\begin{AMS}
68Q32, 46E22, 65N30, 41A25
\end{AMS}

\section{Introduction}

Partial differential equations (PDEs) play a crucial role in modeling a wide
range of processes across diverse fields, including physics, chemistry,
ecology, and biology. Because of their importance, many numerical
methods---including finite element, finite difference, and spectral
methods---have been developed to solve different classes of PDEs
\cite{quarteroni2009numerical}. These traditional methods can become
computationally demanding, especially when high-resolution simulations must be
repeated for varying input parameters.

To address these challenges, data-driven approaches have recently gained
considerable attention \cite{mishra2024numerical}. In particular, operator
learning has emerged as a promising alternative because it can produce
flexible models that are independent of a particular discretization.
Neural-network approaches, in particular, have achieved impressive empirical
performance \cite{alkin2024universal, herde2024poseidon, hao2024dpot}. Despite
these empirical successes, the theoretical understanding of operator learning
remains far from complete \cite{kovachki2024operator, subedi2025operator}. In
particular, deriving PDE-specific guarantees that account for the chosen
algorithm, sampling, and discretization errors remains challenging.

Kernel methods provide an alternative to neural-network approaches, with
recently demonstrated competitive performance in operator learning
\cite{batlle2024kernel}, while
also yielding models that are more amenable to theoretical analysis. However,
existing theoretical results do not generally make explicit how the choice of
kernel should reflect the structure of the target operator in a given problem.
In this work, we take steps toward addressing this gap by considering the
following questions:
\begin{itemize}
\item Can kernel constructions be designed to capture structural and
regularity properties of target operators, with explicit links between the
kernels and the operators?
\item Can widely used kernels, such as Sobolev kernels, be theoretically linked
to specific operators?
\item Can uncertainties arising from sampling and discretization be quantified
explicitly for kernel-based learning algorithms?
\end{itemize}
We address these questions by developing a kernel-based theory for learning
highly regular operators, including analytic operators. The resulting theory
is formulated in a general vector-valued operator-learning setting, while the
results for parametric elliptic PDEs arise as a concrete specialization.

Our
first contribution is to establish explicit connections between analytic
operators and vector-valued reproducing kernel Hilbert spaces (VRKHSs). In particular,
we show that operators whose Taylor coefficients satisfy suitable
geometric decay belong to VRKHSs generated by Szegö kernels. We also show that
the same construction applies to solution maps obtained by inverting
parameter-dependent linear operators. These abstract results are of
independent interest beyond the PDE setting. As a concrete application, we prove that
the solution operator of a parametric elliptic PDE belongs to the
 vector-valued Szegö RKHS.

Our second contribution is a finite-sample analysis of vector-valued kernel
ridge regression in the noiseless setting. For analytic targets belonging to
the Szegö RKHS, we derive exponential convergence in a fractional power of the
sample size, with rates of the form $\exp(-cN^{\alpha})$, $0<\alpha <1$, $c>0$ under an explicit
condition on the kernel parameters. For Sobolev-type kernels, we obtain
polynomial rates that can approach $N^{-\beta}$, $\beta >\frac12$ in the standard $L^2$-error as
the smoothness of the kernel increases. The analysis also applies to a family
of weighted norms, including Sobolev-type norms naturally associated with
analytic RKHSs, and separately quantifies the propagation of bounded
observation noise in the vector-valued setting.

Our third contribution is a specialization of these results to the learning
of parametric elliptic solution operators from finite-element data. The
resulting bounds separate the statistical learning error from the
discretization error and make explicit how the output resolution must scale
with the number of training samples. We complement the theoretical analysis
with experiments in scalar and vector-valued settings. These experiments
illustrate the predicted convergence behavior and compare kernels whose
parameters are either adapted or not adapted to the regularity of the target
operator. Once trained, the kernel surrogate is about an order of magnitude
cheaper to evaluate than a finite-element solve of the same accuracy.

The paper is organized as follows. In
Subsection~\ref{subsec:related_work}, we review related work and situate
our results. Section~\ref{sec:setting} introduces the operator-learning
setting, the parametric elliptic problem, and analytic operators.
Section~\ref{sec:vrkhs_main} presents the relevant vector-valued RKHSs and
kernel constructions, and Section~\ref{sec:vrkhs_analytic} connects them
to analytic operators and parametric elliptic PDEs. Section~\ref{sec:overview} summarizes the main
quantitative conclusions, while Section~\ref{sec:proof_details} develops the
finite-sample error analysis, including the detailed definitions and proofs.
Section~\ref{sec:experimental_overview} presents the experiments. The
appendices collect auxiliary estimates and the proof of the maximal
effective-dimension bound.

\subsection{Related work}\label{subsec:related_work}

The literature relevant to this work spans several interconnected areas:
neural operators, approximation theory for analytic operators, vector-valued
kernel methods, and numerical methods for parametric PDEs. We discuss these
areas separately, emphasizing the connections and distinctions most relevant to
our results.

\paragraph{Neural operators (NOs)}
Over the past few years, a wide range of operator-learning methods based on
neural networks have been developed. Earlier contributions include DeepONets
\cite{lu2021learning} and Fourier neural operators
\cite{li2020fourier}, while more recent work has explored large-scale
Transformer-based architectures aimed at establishing operator-learning
foundation models
\cite{hao2024dpot, herde2024poseidon, alkin2024universal}. Despite their
impressive empirical success, the theoretical understanding of these models
remains rather limited. Existing results are largely focused on universal
approximation guarantees \cite{kovachki2024operator, lanthaler2025nonlocality},
which show that broad classes of operators can be approximated but do not, by
themselves, characterize model size, sample complexity, or optimization.
Other results derive worst-case bounds on the number of parameters and
training samples required to achieve a given accuracy. For standard
neural-network architectures, these bounds are typically exponential for
broad classes of Lipschitz continuous operators, rendering learning infeasible
in general \cite{liu2024deep}. For a comprehensive discussion and an
up-to-date list of references, we refer the reader to the recent surveys
\cite{lanthaler2024error, subedi2025operator}.

\paragraph{Analytic (holomorphic) operators}
More favorable approximation rates can be obtained for classes of operators
with additional regularity. Besides the case of linear operators
\cite{de2023convergence}, a particularly well-studied example is the class of
analytic operators. This regularity assumption is satisfied by many
PDE-related operators, most notably by the parameter-to-solution map of an
elliptic diffusion problem \cite{cohen2015approximation}. Influential
contributions in this direction include
\cite{cohen2010convergence, cohen2011analytic}, with further developments
surveyed in \cite{cohen2015approximation, adcock2022sparse, adcock2024learning,
adcock2024optimal}. Least-squares and compressed-sensing methods turn these
approximation results into practical algorithms with finite-sample guarantees
\cite{chkifa2015discrete, adcock2024efficient}; for a recent empirical comparison,
in which reduced-basis sparse-grid and tensor-train surrogates are substantially
more data-efficient than neural operators for smooth random-field inputs, see
\cite{westermann2026performance}. For neural
networks, by contrast, comparable rates have so far been established only for
the existence of a suitable network \cite{opschoor2022exponential, herrmann2024neural}
or for the minimizer of a training objective
\cite{adcock2024learning, adcock2024optimal}, not for the output of the training
algorithm actually used. The kernel estimator studied here
attains rates of the same type in the analytic case, but selects the hypothesis
space through a kernel rather than an index set and exploits regularization instead of
truncation.

\paragraph{Vector-valued kernels}
We build on kernel methods that explicitly account for the function-valued
nature of the targets. The theoretical foundations for vector-valued kernels
were laid in earlier works
\cite{senkene1973hilbert, carmeli2006vector, carmeli2010vector,
kadri2016operator}, and more recent studies have derived finite-sample bounds
for the associated nonparametric least-squares regression problem
\cite{mollenhauer2022learning, li2024towards, meunier2024optimal}, closely
paralleling results in the scalar-valued setting
\cite{caponnetto2007optimal, smale2007learning, bauer2007regularization}.
Vector-valued kernel methods have recently been shown to achieve strong
empirical performance in operator learning
\cite{batlle2024kernel, batlle2025error, jalalian2025data,
chen2021solving}. In addition, efficient subsampling strategies based on
random features have been proposed
\cite{nelsen2021random, lanthaler2024error, dunbar2025hyperparameter}.
We also note the contribution
\cite{nguyen2024optimal}, which analyzes the generalization properties of
neural operators through neural tangent kernel theory, as well as
\cite{shi2024learning, yang2025learning, yang2025kernel}, which study
stochastic approximation schemes. Despite these advances, existing theoretical
results generally lack an explicit connection to concrete classes of target
operators. Our analysis leverages recent advances in scalar kernel regression
\cite{dommel2024bound, dommel2025approximation} to derive finite-sample
bounds for analytic operators in the noiseless setting, with explicit links
between the target operators and the selected kernels. In the operator-learning
setting, noiseless data correspond to exact, error-free outputs from the
numerical solver.

\paragraph{Parametric (elliptic) PDEs}
Parametric PDEs, in which the solution of a PDE is approximated as a function
of a random or uncertain parameter, are particularly well suited to
operator-learning methods. Many well-established numerical approaches fall
into this category, including polynomial chaos
\cite{ghanem2003stochastic}, stochastic finite elements
\cite{ghanem2003stochastic}, reduced-basis modeling
\cite{hesthaven2016certified, lucia2004reduced}, and sparse polynomial
interpolation techniques \cite{chkifa2013sparse, chkifa2014high}. For
parametric elliptic problems, which serve as our main example, the most
relevant contributions are quadrature-based methods surveyed in
\cite{kuo2016application} and, in particular, the kernel-based framework
introduced in \cite{griebel2017reproducing}. The latter also provides the
foundation for some of the kernel constructions employed here. These
approaches, however, focus on scalar-valued quantities and rely on prescribed
sampling and discretization schemes.

\section{Setting} \label{sec:setting}

We begin by presenting some basic concepts and notation used throughout the
paper: operator learning, parametric elliptic PDEs, and analytic operators.

\subsection{Operator Learning} \label{subsec:op_learn}

Let $\mathcal{X}$ be an input space and let $\mathcal{Y}$ be a real separable
Hilbert space, referred to as the output space. In the parametric PDE setting
considered below, the input space is
$\mathcal{X}=[-1,1]^P$. The goal is to learn an estimate $\hat{F}$ of a target
operator $F_*\colon\mathcal{X}\to\mathcal{Y}$ from samples
$z=(x_i,y_i)_{i=1}^N$, where
\begin{equation}\label{eq:xiyi}
y_i = F_*(x_i) + \varepsilon_i,
\qquad i=1,\dots,N.
\end{equation}
The inputs $x_i$ are drawn independently and identically distributed
according to a probability measure $\mu$ on $\mathcal{X}$. When $\mathcal{X}=[-1,1]^P$, we might also impose the following assumption on the sampling distribution.
\begin{assumption}[Bounded density]\label{as:sampling}
Set $\mathcal{X}=[-1,1]^P$. The measure $\mu$ admits a density with respect to the Lebesgue measure, also denoted by $\mu$, satisfying
\begin{equation}
\label{bound_dens}
0<c_\mu\leq\mu(x)\leq C_\mu,
\qquad x\in\mathcal{X}.
\end{equation}
\end{assumption}
We impose the following assumption on the observation noise.
\begin{assumption}[Observation noise]\label{as:noise}
The noise variables $\varepsilon_i\in\mathcal{Y}$ are independent and
identically distributed, independent of the inputs, centered, and almost
surely bounded: for some $L>0$,
\[
\EXP[\varepsilon_i]=0,
\qquad
\norm{\varepsilon_i}_{\mathcal{Y}}\leq L
\quad\text{almost surely}.
\]
\end{assumption}
The noiseless case corresponds to $\varepsilon_i=0$, so that
$y_i=F_*(x_i)$. When $\mathcal{Y}=\mathbb{R}$, the framework reduces to
standard regression.

The accuracy of an estimate $\hat F$ can be measured by the mean-squared error
\begin{equation}\label{eq:l2_error}
\norm{F_*-\hat{F}}_{L^2(\mathcal{X},\mathcal{Y})}^2
=
\EXP_{x\sim\mu}
\left[\norm{F_*(x)-\hat{F}(x)}_{\mathcal{Y}}^2\right],
\end{equation}
but we will also consider more general error metrics.
Given the data and a hypothesis class
$\mathcal{H}\subseteq\mathcal{Y}^{\mathcal{X}}$, a classical approach is to
define $\hat{F}$ as the solution of a regularized empirical risk minimization
problem,
\begin{align} \label{gen_erm}
\hat{F}
=
\underset{F\in\mathcal{H}}{\arg\min}\;
\frac{1}{N}\sum_{i=1}^N
\left\|y_i-F(x_i)\right\|_{\mathcal{Y}}^2
+\lambda\Lambda(F),
\end{align}
where $\Lambda\colon\mathcal{H}\to[0,\infty)$ is a regularization functional
and $\lambda>0$ is the regularization parameter. In the noiseless case, we
denote the corresponding solution by $\bar{F}$.

In what follows, we study how different choices of the hypothesis class affect
finite-sample error bounds, with particular emphasis on the case where the
unknown operator is analytic. Our motivating applications concern the learning
of solution operators for parametric PDEs, which we describe next.

\subsection{Parametric PDEs}\label{sub:parametricPDE}

In the PDE setting, $\mathcal{Y}$ is a Hilbert space of functions defined on a
bounded domain $\Omega\subseteq\mathbb{R}^d$. For the elliptic problem below,
we assume that $H_0^1(\Omega)$ is continuously embedded in the chosen output
space $\mathcal{Y}$. We consider boundary-value problems of the form
\begin{align*}
\mathcal{L}_x y &= f \qquad \text{in } \Omega,
\end{align*}
subject to suitable boundary conditions on $\partial\Omega$, where
$\mathcal{L}_x$ is a differential operator parametrized by $x$, $f$ is a
fixed source term, and the solution $y$ belongs to $\mathcal{Y}$.

Our main example is a parametric elliptic PDE. For a given parameter
$x\in\mathcal{X}$, let $y$ be a solution of the boundary-value problem
\begin{equation}\label{eq:elliptic}
\begin{aligned}
-\nabla\cdot\bigl(a(x,s)\nabla y(s)\bigr) &= f(s),
&& s\in\Omega,\\
y(s)&=0,
&& s\in\partial\Omega.
\end{aligned}
\end{equation}
Then, whenever the problem is well posed, we let
$F_*^{\mathrm{ell}}\colon\mathcal{X}\to\mathcal{Y}$ denote the solution
operator, given by $F_*^{\mathrm{ell}}(x)=y$ for each $x$.
We model the diffusion coefficient using the finite-dimensional affine
parametrization
\begin{align} \label{eq:input_coeff}
a(x,s)=\phi_0(s)+\sum_{p=1}^P x_p\phi_p(s).
\end{align}
Here, $x=(x_1,\ldots,x_P)\in\mathcal{X}=[-1,1]^P$, and
$\phi_p\in L^\infty(\Omega)$, for $p=0,\ldots,P$, are fixed functions.

We impose the following standard regularity and ellipticity assumptions on
\eqref{eq:elliptic}.

\begin{assumption} \label{as:pde}
Let $\mathcal{X}=[-1,1]^P$, with $\mu$ satisfying
\eqref{bound_dens}. Moreover, for $d=1,2,3$, let
$\Omega\subset\mathbb{R}^d$ be bounded and convex, with polygonal or
polyhedral boundary as appropriate. Let $f\in L^2(\Omega)$, and let the
coefficient be given by \eqref{eq:input_coeff} and satisfy the uniform
ellipticity condition
\begin{align} \label{def:unif_ell}
a_{\mathrm{min}}\leq a(x,s)\leq a_{\mathrm{max}},
\qquad s\in\Omega,\ x\in\mathcal{X},
\end{align}
for some $0<a_{\mathrm{min}}\leq a_{\mathrm{max}}<\infty$. Moreover,
$\phi_0,\phi_1,\ldots,\phi_P$ are Lipschitz continuous on $\overline{\Omega}$.
\end{assumption}

Under Assumption~\ref{as:pde}, for every $x\in\mathcal{X}$, problem
\eqref{eq:elliptic} admits a unique weak solution
$F_*^{\mathrm{ell}}(x)\in H_0^1(\Omega)$ by the Lax--Milgram theorem
\cite{evans2022partial}. By the continuous embedding
$H_0^1(\Omega)\hookrightarrow\mathcal{Y}$, this solution also belongs to
$\mathcal{Y}$. By elliptic regularity for convex polygonal or polyhedral
domains, this solution in fact belongs to $H^2(\Omega)$
\cite{grisvard2011elliptic}. In Section~\ref{sec:pde}, we discuss further
details on \eqref{eq:elliptic} and its numerical discretization.

Although, for fixed $a$, the differential operator
$\mathcal{L}_x(y)=-\nabla\cdot\bigl(a(x,\cdot)\nabla y\bigr)$
is linear in $y$ and, for fixed $y$, linear in $a$, the solution operator
$F_*^{\mathrm{ell}}$ is inherently nonlinear, since it involves inverting
$\mathcal{L}_x$.

The solution operators arising in many applications, including the elliptic
problem above, satisfy strong regularity conditions
\cite{cohen2010convergence, cohen2011analytic}. These conditions can imply
analyticity on $\mathcal{X}$, often through a holomorphic extension to a
complex neighborhood of $\mathcal{X}$ in $\mathbb{C}^P$; see, for example,
\cite{adcock2024learning}. We discuss this idea next.

\subsection{Analytic (holomorphic) operators}

A multi-index $\nu$ is an element of $\mathbb{N}^P$, where
$\mathbb{N}=\{0,1,2,\ldots\}$. For $\nu\in \mathbb{N}^P$ and $x\in\mathbb{R}^P$ we use the standard notation
\[
\abs{\nu}=\sum_{p=1}^P\nu_p,
\qquad
\nu!=\nu_1!\cdots\nu_P!,\qquad
x^\nu=\prod_{p=1}^P x_p^{\nu_p}.
\]
The following definition is adapted from
\cite{cohen2010convergence, cohen2011analytic}.

\begin{definition}[Analytic operators]\label{def:analytic_op}
Let $\mathcal{X}=[-1,1]^P$. An operator
$F_*\colon\mathcal{X}\to\mathcal{Y}$ is called \textit{analytic} if
\begin{align} \label{eq:analytic_op}
F_*(x)=\sum_{\nu\in\mathbb{N}^P}y_\nu x^\nu,
\qquad x\in\mathcal{X},
\end{align}
where $y_\nu\in\mathcal{Y}$ and the series converges absolutely in
$\mathcal{Y}$ for every $x\in\mathcal{X}$.
\end{definition}

Definition~\ref{def:analytic_op} does not imply a holomorphic extension beyond $\X$; such an extension holds for the elements of the reproducing kernel Hilbert space defined by  Szegö kernels  considered below. The above definition gives a power-series representation of analytic operators,
but does not yet specify how this regularity can be exploited in learning. A
fundamental question is therefore how to choose an appropriate hypothesis
space $\Hs$ of operators, together with an associated learning algorithm, for
approximating analytic operators $F_*$ such as $F_*^{\mathrm{ell}}$. Ideally,
one would choose $\Hs$ so that $F_*\in\Hs$. In the following sections, we
address this question in the context of reproducing kernel Hilbert spaces by
constructing suitable kernels and later analyzing the corresponding learning
algorithms.

\section{Kernel spaces for analytic operators and parametric PDEs}
\label{sec:vrkhs_main}

In this section, we introduce the hypothesis spaces used throughout the paper,
namely, reproducing kernel Hilbert spaces. We review
standard definitions, properties, and examples of scalar and vector-valued
kernels. Their connection to the analytic operators considered in this
work is established in Section~\ref{sec:vrkhs_analytic}.

\subsection{Vector reproducing kernel Hilbert spaces}
\label{subsec:kernel_def}

Let $\mathcal{X}$ be a set, and let $\mathcal{Y}$ be a real separable Hilbert
space. We denote by $\mathcal{L}(\mathcal{Y})$ the space of bounded linear
operators on $\mathcal{Y}$. A vector-valued reproducing kernel is a map
$K\colon\mathcal{X}\times\mathcal{X}\to\mathcal{L}(\mathcal{Y})$ satisfying
$K(x,x')=K(x',x)^*$ and, for every $M\in\mathbb{N}$,
$x_1,\ldots,x_M\in\mathcal{X}$, and $y_1,\ldots,y_M\in\mathcal{Y}$,
\[
\sum_{i,j=1}^M
\innerpro{y_i,K(x_i,x_j)y_j}_{\mathcal{Y}}
\geq 0.
\]
The corresponding vector-valued reproducing kernel Hilbert space (VRKHS)
$\Hs_K$ is the completion of
$\operatorname{span}\{K(\cdot,x)y:x\in\mathcal{X},\,y\in\mathcal{Y}\}$
with respect to the inner product determined by
$\innerpro{K(\cdot,x)y,K(\cdot,x')y'}_{\Hs_K}
=
\innerpro{y,K(x,x')y'}_{\mathcal{Y}}$.
The space $\Hs_K$ satisfies the reproducing property
\begin{align}\label{eq:reproducing_prop_vector}
\innerpro{F,K(\cdot,x)y}_{\Hs_K}
=
\innerpro{F(x),y}_{\Y},
\qquad
F\in\Hs_K,\quad x\in\X,\quad y\in\Y.
\end{align}
The space $\Hs_K$ is uniquely determined by $K$, and conversely $K$ is
uniquely determined by the reproducing property. When the kernel is clear
from context, we omit the subscript $K$.

The theory of reproducing kernels in vector-valued settings goes back at
least to Schwartz~\cite{schwartz1964sous}. Subsequent developments of
vector-valued reproducing kernel Hilbert spaces can be found, for example, in
\cite{carmeli2006vector,carmeli2010vector}; see also
\cite{senkene1973hilbert,kadri2016operator}.

Next, we make the  first step towards designing task-specific kernels.

\subsection{From scalar to separable kernels}

Separable kernels provide a simple and flexible way to extend familiar
scalar kernel constructions to vector-valued outputs. We first briefly review
some properties of scalar kernels and related quantities.

\paragraph{Scalar kernels}

A scalar kernel $k\colon\X\times\X\to\Real$ defines a scalar-valued RKHS
$\Hs_k$, corresponding to the special case $\Y=\Real$. In this case, the
reproducing property becomes
$f(x)=\innerpro{f,k(x,\cdot)}_{\Hs_k}$ for $f\in\Hs_k$ and $x\in\X$.
Let $\varphi(x)=k(x,\cdot)\in\Hs_k$.
Throughout the paper we consider the following boundedness assumption.

\begin{assumption}[Bounded kernel]\label{as:bounded_kernel}
The  kernel  $k\colon\X\times\X\to\Real$ satisfies
$\sup_{x\in\mathcal{X}} k(x,x)\leq\kappa^2$
for some $\kappa>0$.
\end{assumption}

The covariance operator
$\Sigma\colon\Hs_k\to\Hs_k$ is defined by
\[
\Sigma=\int_{\X}\varphi(x)\otimes\varphi(x)\,\mathrm{d}\mu(x),
\]
and its empirical counterpart
$\hat{\Sigma}\colon\Hs_k\to\Hs_k$ is given by
\[
\hat{\Sigma}
=
\frac{1}{N}\sum_{i=1}^N
\varphi(x_i)\otimes\varphi(x_i).
\]
Under Assumption~\ref{as:bounded_kernel}, $\Sigma$ is positive and trace class, and hence compact.

Let $(\sigma_l,f_l)_{l\geq1}$ be the positive eigensystem of $\Sigma$, with
$\Sigma f_l=\sigma_l f_l$, $\norm{f_l}_{\Hs_k}=1$, and
$\sigma_l\searrow0$. Then
$\sigma_1=\norm{\Sigma}_{\Hs_k\to\Hs_k}\leq\kappa^2$. For every $l\geq1$,
the function $e_l=\sigma_l^{-1/2}f_l$ belongs to
$L^2(\X,\mu)$, and $(e_l)_{l\geq1}$ is an orthonormal system in $L^2(\X,\mu)$.

Several quantities admit explicit expressions in terms of the covariance
operators. In particular, the $L^2$ norm of every $f\in\Hs_k$ satisfies
\begin{equation}\label{scal_norm_eq}
\norm{f}_{L^2(\X,\Real)}
=
\norm{\Sigma^{1/2}f}_{\Hs_k}.
\end{equation}
Moreover, the empirical risk minimization~\eqref{gen_erm} in RKHS, commonly
called kernel ridge regression, is
\begin{align}\label{scal_krr}
\hat{f}
=
\underset{f\in\Hs_k}{\arg\min}\;
\frac{1}{N}\sum_{i=1}^N
\abs{y_i-f(x_i)}^2
+\lambda\norm{f}_{\Hs_k}^2.
\end{align}
Its unique solution admits the explicit representation
\begin{equation}\label{eq:representer_noise}
\hat{f}= \hat{\Sigma}_{\lambda}^{-1}g_z,
\end{equation}
where $\hat{\Sigma}_{\lambda}=\hat{\Sigma}+\lambda\I_{\Hs_k}$ and
$g_z=\frac{1}{N}\sum_{i=1}^N\varphi(x_i)y_i\in\Hs_k$.
For noiseless data, if $f_*\in\Hs_k$, the corresponding estimator is
\begin{equation}\label{eq:representer_noiseless}
\bar{f}
=
\hat{\Sigma}_{\lambda}^{-1}\hat{\Sigma}f_*.
\end{equation}
As shown below, these formulas extend to separable kernels and are
instrumental in the subsequent error analysis.

\paragraph{Separable kernels}

Separable kernels \cite{alvarez2012kernels} have the form
\begin{equation}\label{sep_ker}
K(x,x')=k(x,x')\I_{\Y},
\end{equation}
where $k$ is a scalar-valued kernel and $\I_{\Y}$ is the identity operator on
$\Y$. In the following we always assume the scalar kernel to satisfy Assumption~\ref{as:bounded_kernel}.

Separable kernels provide a simple way to extend scalar kernel constructions and analysis to
vector-valued outputs. Indeed, let $\Hs_k$ denote the scalar RKHS associated
with $k$, and let $\varphi(x)=k(x,\cdot)$. The VRKHS associated with a
separable kernel admits the tensor-product representation
\begin{align}\label{eq:sep_isom}
\Hs_K
\cong
\Y\otimes\Hs_k
\cong
S_2(\Y,\Hs_k),
\end{align}
where $S_2(\Y,\Hs_k)$ denotes the space of Hilbert--Schmidt operators from $\Y$ to $\Hs_k$; see, for example,
\cite[Theorem~12.6.1]{aubin2011applied}.
In particular, for every $F\in\Hs_K$, there exists a unique
$C\in S_2(\Y,\Hs_k)$ such that
$F(x)=C^*\varphi(x)$ for $x\in\X$.
This representation is the key property allowing us to derive explicit
expressions in terms of covariance operators, analogous to those in the
scalar case. In particular, the $L^2$ norm of $F\in\Hs_K$ satisfies
\begin{equation}\label{vec_norm_eq}
\norm{F}_{L^2(\X,\Y)}
=
\norm{\Sigma^{1/2}C}_{S_2(\Y,\Hs_k)},
\end{equation}
see \cite[Lemma~2]{meunier2024optimal}. This is the vector-valued analogue
of the scalar identity~\eqref{scal_norm_eq}. Moreover, vector-valued kernel
ridge regression can be written as
\begin{align}\label{vec_krr}
\hat{F}
=
\underset{F\in\Hs_K}{\arg\min}\;
\frac{1}{N}\sum_{i=1}^N
\norm{y_i-F(x_i)}_{\Y}^2
+\lambda\norm{F}_{\Hs_K}^2,
\end{align}
and its unique solution is given by
$\hat{F}(x)=\hat{C}^*\varphi(x)$, where
$\hat{C}=\hat{\Sigma}_{\lambda}^{-1}G_z$,
and
\[
G_z
=
\frac{1}{N}\sum_{i=1}^N
\varphi(x_i)\otimes y_i
\in S_2(\Y,\Hs_k),
\]
where we use the following convention for the tensor product throughout the text:
$(h\otimes y)v=\innerpro{y,v}_{\Y}h$.
For noiseless data, if $F_*(x)=C_*^*\varphi(x)$, the corresponding estimator
is $\bar{F}(x)=\bar{C}^*\varphi(x)$ with
$\bar{C}=\hat{\Sigma}_{\lambda}^{-1}\hat{\Sigma}C_*$.
These expressions enable the analysis of vector-valued problems through minor
extensions of the scalar-valued theory. Since our analysis of analytic operators and parametric PDEs uses separable
kernels, we next describe the relevant examples of scalar kernels from which
they can be constructed.

\subsection{Examples of kernels}

We consider three main classes of scalar kernels, each of which induces a
corresponding separable kernel. The first two examples are defined on
$\X=[-1,1]^P$, while the last will also be considered on more general bounded
domains.

\paragraph{Szegö and power-series kernels}

The Szegö kernel is the most relevant example for our study and is defined by
\begin{equation}\label{eq:def_Szeg}
k_{\mathrm{Szeg\ddot{o}}}(x,x')
=
\prod_{p=1}^P\frac{1}{1-c_px_px'_p}
=
\sum_{\nu\in\mathbb{N}^P}
c^\nu x^\nu (x')^\nu,
\end{equation}
where $0<c_p<1$ for $1\leq p\leq P$, and $c^\nu=\prod_{p=1}^Pc_p^{\nu_p}$. More generally, a power-series kernel is
given by
\begin{equation}\label{def:power_kernel}
k_{\mathrm{pow}}(x,x')
=
\sum_{\nu\in\mathbb{N}^P}
\frac{w_\nu}{(\nu!)^2}x^\nu(x')^\nu,
\qquad
x,x'\in[-1,1]^P,
\end{equation}
where $(w_\nu)_{\nu\in\mathbb{N}^P}$ is a sequence of positive weights
satisfying
\(
\sum_{\nu\in\mathbb{N}^P}
\frac{w_\nu}{(\nu!)^2}<\infty.
\)
The Szegö kernel is a special case of this family, with
$w_\nu=(\nu!)^2c^\nu$. The RKHS associated with a power-series kernel admits
the characterization \cite{zwicknagl2009power}
\begin{align}\label{eq:pow_series_rkhs}
\Hs_{k_{\mathrm{pow}}}
=
\left\{
f(x)=\sum_{\nu\in\mathbb{N}^P}f_\nu x^\nu:
\sum_{\nu\in\mathbb{N}^P}
\frac{(\nu!)^2}{w_\nu}f_\nu^2<\infty
\right\},
\end{align}
with inner product
\begin{align*}
\innerpro{f,g}_{\Hs_{k_{\mathrm{pow}}}}
=
\sum_{\nu\in\mathbb{N}^P}
\frac{(\nu!)^2}{w_\nu}f_\nu g_\nu.
\end{align*}
In particular, for the Szegö kernel with
$P=1$ we have
\begin{align}\label{eq:szeg_rkhs_1d}
\Hs_{k_{\mathrm{Szeg\ddot{o}}}}
=
\left\{
g(x)=\sum_{l\geq0}g_lx^l:
\norm{g}_{\Hs_{k_{\mathrm{Szeg\ddot{o}}}}}^2
=
\sum_{l\geq0}\frac{1}{c^l}g_l^2<\infty
\right\}.
\end{align}

The corresponding vector-valued space admits an analogous
characterization.

\begin{lemma}
\label{lem:power_series_vrkhs}
Let $\X=[-1,1]^P$, let $k_{\mathrm{pow}}$ be a power-series
kernel as in~\eqref{def:power_kernel}, and set
$K_{\mathrm{pow}}=k_{\mathrm{pow}}\I_{\Y}$. Then
\begin{equation}\label{eq:pow_series_vrkhs}
\Hs_{K_{\mathrm{pow}}}
=
\left\{
F(x)=\sum_{\nu\in\mathbb{N}^P}y_\nu x^\nu:
y_\nu\in\Y,\;\norm{F}_{\Hs_{K_{\mathrm{pow}}}}^2=\sum_{\nu\in\mathbb{N}^P}
\frac{(\nu!)^2}{w_\nu}\norm{y_\nu}_{\Y}^2<\infty
\right\}.
\end{equation}
Every such series converges absolutely
and uniformly on $\X$.
\end{lemma}

\begin{proof}
By~\eqref{eq:pow_series_rkhs} and the associated inner product,
the functions
$\phi_\nu(x)=\sqrt{w_\nu}\,x^\nu/\nu!$, $\nu\in\mathbb{N}^P$,
form an orthonormal basis of $\Hs_{k_{\mathrm{pow}}}$.
Let $(h_l)_l$ be an orthonormal basis of $\Y$.
The tensor-product identification~\eqref{eq:sep_isom}
then shows that
\(
F_{\nu,l}(x)=\phi_\nu(x)h_l
\)
form an orthonormal basis of $\Hs_{K_{\mathrm{pow}}}$.
Consequently, this space consists of the expansions
$F=\sum_{\nu,l}b_{\nu,l}F_{\nu,l}$ with $\sum_{\nu,l}|b_{\nu,l}|^2<\infty$,
with squared norm $\sum_{\nu,l}|b_{\nu,l}|^2$.
The correspondence
\[
y_\nu=\frac{\sqrt{w_\nu}}{\nu!}\sum_l b_{\nu,l}h_l,
\qquad
b_{\nu,l}=\frac{\nu!}{\sqrt{w_\nu}}
\innerpro{y_\nu,h_l}_{\Y},
\]
and Parseval's identity give
$\sum_{\nu}(\nu!)^2w_\nu^{-1}\norm{y_\nu}_{\Y}^2=\sum_{\nu,l}|b_{\nu,l}|^2$.
This proves the characterization and the norm identity.
Finally, the Cauchy--Schwarz inequality and the summability
of $w_\nu/(\nu!)^2$ imply
$\sum_\nu\norm{y_\nu}_{\Y}<\infty$.
Since $|x^\nu|\leq1$ for every  $x\in\X$, the power series converges
absolutely and uniformly.
\end{proof}

\paragraph{Gaussian kernels}

The Gaussian, or radial basis function (RBF), kernel
$k_{\text{Gauss}}$ is defined by
\begin{equation} \label{eq:gauss}
k_{\text{Gauss}}(x,x')
=
\exp\left(-\bar s^2\norm{x-x'}^2\right),
\end{equation}
where $\bar s>0$ is a scaling parameter. Gaussian kernels are perhaps the most
common analytic kernels used in practice. Their RKHS, however, has a
markedly different structure from that of the Szegö kernel. For simplicity,
we restrict to the case $P=1$. The Gaussian RKHS has the characterization
\cite{minh2010some,steinwart2008support}
\begin{align}\label{eq:gauss_rkhs_1d}
\Hs_{k_{\text{Gauss}}}
=
\left\{
g(x)=e^{-\bar s^2x^2}\sum_{l\geq0}g_lx^l:
\norm{g}_{\Hs_{k_{\text{Gauss}}}}^2
=
\sum_{l\geq0}
\left(\frac{1}{2\bar s^2}\right)^l l!g_l^2<\infty
\right\}.
\end{align}
The following result compares this space with the Szegö RKHS characterized
in~\eqref{eq:szeg_rkhs_1d}.

\begin{lemma}
\label{lem:gauss_szeg}
For $P=1$, $\Hs_{k_{\text{Gauss}}}\subset\Hs_{k_{\mathrm{Szeg\ddot{o}}}}$.
Moreover, the function $f(x)=(1-cx)^{-1}$ belongs to
$\Hs_{k_{\mathrm{Szeg\ddot{o}}}}$ for $0<c<1$, but not to
$\Hs_{k_{\text{Gauss}}}$.
\end{lemma}

\begin{proof}
Let $g\in\Hs_{k_{\text{Gauss}}}$ and write
$g(x)=e^{-\bar s^2x^2}\sum_{l\geq0}g_lx^l$. Then
$g(x)=\sum_{l\geq0}h_lx^l$, where
$h_l=\sum_{k=0}^l a_{l-k}g_k$, with
$a_{2l}=(-1)^l \bar{s}^{2l}/l!$ and $a_{2l+1}=0$.
Set
$\widetilde a_l=a_lc^{-l/2}$,
$\widetilde g_l=g_lc^{-l/2}$, and
$\widetilde h_l=h_lc^{-l/2}$. Then
$\widetilde h=\widetilde a*\widetilde g$. The sequence
$(\widetilde a_l)_{l\geq0}$ belongs to $\ell^1$, while
$(\widetilde g_l)_{l\geq0}$ belongs to $\ell^2$: indeed, for sufficiently
large $l$, $c^{-l}\leq(2\bar s^2)^{-l}l!$, and the latter sequence is
square-summable by the definition of $\Hs_{k_{\text{Gauss}}}$. Young's
inequality therefore gives $\widetilde h\in\ell^2$, which proves that
$g\in\Hs_{k_{\mathrm{Szeg\ddot{o}}}}$.

Now let $f(x)=(1-cx)^{-1}=\sum_{l\geq0}c^lx^l$. Writing
$f(x)=e^{-\bar s^2x^2}\sum_{l\geq0}h_lx^l$, we have
$h_l=\sum_{k=0}^l b_{l-k}c^k$, where
$b_{2l}=\bar s^{2l}/l!$ and $b_{2l+1}=0$. Since $b_0=1$, we have
$h_l\geq c^l$. Hence
\[
\norm{f}_{\Hs_{k_{\text{Gauss}}}}^2
=
\sum_{l\geq0}(2\bar s^2)^{-l}l!h_l^2
\geq
\sum_{l\geq0}
\left(\frac{c^2}{2\bar s^2}\right)^l l!
=
\infty.
\]
Therefore, $f\notin\Hs_{k_{\text{Gauss}}}$.
\end{proof}

\paragraph{Sobolev kernels}

We finally consider Sobolev-type kernels. Let $\X\subset\mathbb{R}^P$ be
bounded with Lipschitz boundary, and assume that $\mu$ has a density
satisfying $0<c_\mu\leq\mu(x)\leq C_\mu$ for $x\in\X$. Let
$\bar r:=r-P/2>0$, for some $r>P/2$.

Sobolev-type RKHSs can, for instance, be constructed using Matérn kernels:
\begin{align}\label{eq:matern}
k_{\mathrm{Sob}}(x,x')
=
\frac{1}{\Gamma(\bar r)2^{\bar r-1}}
\bigl(\sqrt{2\bar r}\,\bar s\norm{x-x'}\bigr)^{\bar r}
\mathcal{K}_{\bar r}
\bigl(\sqrt{2\bar r}\,\bar s\norm{x-x'}\bigr),
\end{align}
where $\bar r$ determines the regularity, $\bar s>0$ is a scaling parameter,
$\Gamma$ denotes the Gamma function, and $\mathcal{K}_{\bar r}$ is the
modified Bessel function of the second kind; see, for example,
\dlmf{5.2.1} and \dlmf{10.25}, respectively.

We write $H^m(\X)$, $m\in\mathbb{N}$, for the classical Sobolev space.
The fractional Sobolev space $H^r(\X)$ is obtained, up to equivalent norms, by
real interpolation between $L^2(\X)$ and $H^m(\X)$ for any integer $m>r$; for
bounded Lipschitz domains, it is independent of $m$ and coincides with the
Sobolev--Slobodeckij space \cite{adams2003sobolev}. The properties of
Sobolev-type RKHSs needed below are collected in the following lemma; see
\cite{adams2003sobolev,wendland2004scattered,tuo2016theoretical,fischer2020sobolev,zhang2024optimality}.
These references assume a smooth boundary; the extension to Lipschitz domains
follows by means of a bounded extension operator
\cite[Chapter~VI]{stein1970singular} and the corresponding embedding
estimates \cite[Theorem~3.3.4]{edmunds1996function}.

\begin{lemma}
\label{lem:sob_char}
Under the assumptions above:

\begin{enumerate}[label=(\roman*)]
\item The functions in $H^r(\X)$ are continuous.

\item For the Matérn kernel~\eqref{eq:matern}, the RKHS
$\Hs_{k_{\mathrm{Sob}}}$ is norm-equivalent to $H^r(\X)$.

\item If $P/(2r)<a\leq1$, then there exist constants
$C_a,c_r,C_r>0$ such that, for every $l\geq1$ and $x\in\X$,
\begin{align}
c_r l^{-2r/P}
\leq \sigma_l
&\leq C_r l^{-2r/P},
\label{eq:evaules_decay}\\
\sum_{l\geq1}\sigma_l^a e_l(x)^2
&\leq C_a.
\label{eq:eigenfcts}
\end{align}
\end{enumerate}
\end{lemma}

All three scalar kernels considered above satisfy
Assumption~\ref{as:bounded_kernel}. The maximal effective dimension and its
corresponding estimate are introduced later, together with the learning
bounds.

With these preliminaries, we are now in a position to discuss how to design
kernels for representing analytic operators and solution operators of
parametric PDEs.

\section{\texorpdfstring{VRKHS for analytic operators and parametric PDEs}{VRKHS for analytic operators and parametric PDEs}}
\label{sec:vrkhs_analytic}

We now ask whether the class of analytic operators defined in
\eqref{eq:analytic_op} belongs to a VRKHS generated by a separable
kernel of the form~\eqref{sep_ker}. The following theorem gives a general
criterion for power-series kernels and, as a special case, establishes the
connection with the Szegö kernel.

\begin{theorem}[VRKHS embedding for analytic operators]
\label{thm:main_kernel_op}
Let $F_*$ be an analytic operator as in Definition~\ref{def:analytic_op},
and let $k_{\mathrm{pow}}$ be a power-series kernel of the form
\eqref{def:power_kernel}, with associated weights
$(w_\nu)_{\nu\in\mathbb{N}^P}$. If
\begin{align}\label{eq:taylor_cond}
\sum_{\nu\in\mathbb{N}^P}
\frac{(\nu!)^2\norm{y_\nu}_{\Y}^2}{w_\nu}
<\infty,
\end{align}
then
\(
F_*\in\Hs_{K_{\mathrm{pow}}},
\)
where
$K_{\mathrm{pow}}=k_{\mathrm{pow}}\I_{\Y}$.
In particular, suppose that
\begin{align}\label{eq:taylor_est}
\norm{y_\nu}_{\Y}\leq C\rho^\nu,
\qquad
\rho^\nu:=\prod_{p=1}^P\rho_p^{\nu_p},
\end{align}
for some $C>0$ and $0\leq \rho_p<\sqrt{c_p}$, $1\leq p\leq P$. Then
$$
F_*\in\Hs_{K_{\mathrm{Szeg\ddot{o}}}},
$$
where $K_{\mathrm{Szeg\ddot{o}}}=k_{\mathrm{Szeg\ddot{o}}}\I_{\Y}$ and $k_{\mathrm{Szeg\ddot{o}}}$ is given in \eqref{eq:def_Szeg}.
\end{theorem}

\begin{proof}
By Lemma~\ref{lem:power_series_vrkhs},
condition~\eqref{eq:taylor_cond} implies
$F_*\in\Hs_{K_{\mathrm{pow}}}$.

For the Szegö kernel, the weights are
$w_\nu=(\nu!)^2c^\nu$. Using~\eqref{eq:taylor_est}, we obtain
\[
\sum_{\nu\in\mathbb{N}^P}
\frac{\norm{y_\nu}_{\Y}^2}{c^\nu}
\leq
C^2\sum_{\nu\in\mathbb{N}^P}
\prod_{p=1}^P
\left(\frac{\rho_p^2}{c_p}\right)^{\nu_p}=
C^2\prod_{p=1}^P
\frac{1}{1-\rho_p^2/c_p}
<\infty,
\]
since $\rho_p^2<c_p$ for every $p$.
Another application of Lemma~\ref{lem:power_series_vrkhs}
therefore yields
$F_*\in\Hs_{K_{\mathrm{Szeg\ddot{o}}}}$.
\end{proof}

The Szegö kernel is also naturally connected to affine inverses of linear
operators.

\begin{proposition}[VRKHS for affine inverses of linear operators]
\label{pro:inversion_abstract}
Let \(\mathcal X=[-1,1]^P\), and let \(\mathcal Y\) be a real separable Hilbert space. Let
$S_0\in\mathcal{L}(\Y)$ be boundedly invertible and let
$S_p\in\mathcal{L}(\Y)$ for $1\leq p\leq P$, and let $0<c_p<1$ be the parameters of the Szegö kernel~\eqref{eq:def_Szeg}. Assume that
\begin{align}\label{eq:summ_condition_inversion}
\sum_{p=1}^P\frac{\rho_p}{\sqrt{c_p}}<1,\qquad \rho_p:=
\norm{S_0^{-1}S_p}_{\mathcal{L}(\Y)}.
\end{align}
For
\[
S(x)=S_0+\sum_{p=1}^P x_pS_p\in\mathcal{L}(\Y),
\qquad x\in\X,
\]
the operator $S(x)$ is invertible for every $x\in\X$. Moreover, for every
$g\in\Y$, the mapping
$F_*\colon\X\to\Y$, $F_*(x)=(S(x))^{-1}g$,
belongs to $\Hs_{K_{\mathrm{Szeg\ddot{o}}}}$.
\end{proposition}

\begin{proof}
Set
$A_p=S_0^{-1}S_p$, $b=S_0^{-1}g$, and $A(x)=\sum_{p=1}^P x_pA_p$.
Then
$S(x)=S_0(\I_{\Y}+A(x))$.
Moreover,
$\norm{A(x)}_{\mathcal{L}(\Y)}\leq\sum_{p=1}^P\rho_p\leq\sum_{p=1}^P\rho_p/\sqrt{c_p}<1$.
Thus $\I_{\Y}+A(x)$, and hence $S(x)$, is invertible for every
$x\in\X$.

For a word $w=w_1\cdots w_m$ with letters
$w_j\in\{1,\ldots,P\}$, write
$A^w=A_{w_1}\cdots A_{w_m}$.
For $\nu\in\mathbb{N}^P$, let $W(\nu)$ denote the set of words of length
$|\nu|$ in which the letter $p$ appears exactly $\nu_p$ times, and define
$\overline{A}^{\,\nu}=\sum_{w\in W(\nu)}A^w$.
The empty word corresponds to $\nu=0$ and
$\overline{A}^{\,0}=\I_{\Y}$. By the multinomial expansion,
$A(x)^m=\sum_{|\nu|=m}x^\nu\overline{A}^{\,\nu}$.
Therefore, the Neumann series gives
$$
F_*(x)
=
(\I_{\Y}+A(x))^{-1}b=
\sum_{m=0}^\infty(-1)^mA(x)^m b
=
\sum_{\nu\in\mathbb{N}^P}
x^\nu y_\nu,
$$
where
\(
y_\nu=(-1)^{|\nu|}\overline{A}^{\,\nu}b
\).
Since
\(
|W(\nu)|=\frac{|\nu|!}{\nu!},
\)
and $\rho^\nu:=\prod_{p=1}^P\rho_p^{\nu_p}$, we have
\(
\norm{\overline{A}^{\,\nu}}_{\mathcal{L}(\Y)}
\leq
\frac{|\nu|!}{\nu!}\rho^\nu.
\)
Consequently,
$$
\sum_{\nu\in\mathbb{N}^P}
\frac{\norm{y_\nu}_{\Y}^{2}}{c^\nu}
\leq
\norm{b}_{\Y}^{2}
\left(
\sum_{\nu\in\mathbb{N}^P}
\frac{|\nu|!}{\nu!}
\frac{\rho^\nu}{\sqrt{c^\nu}}
\right)^2
=
\norm{b}_{\Y}^{2}
\left(
\sum_{m=0}^\infty
\left(
\sum_{p=1}^P
\frac{\rho_p}{\sqrt{c_p}}
\right)^m
\right)^2
<\infty,
$$
where the equality follows from the multinomial theorem and the final
inequality from~\eqref{eq:summ_condition_inversion}. Since the Szegö weights are $w_\nu=(\nu!)^2c^\nu$, this is
condition~\eqref{eq:taylor_cond}, which by Lemma~\ref{lem:power_series_vrkhs}
also ensures absolute convergence of the series; the first part of
Theorem~\ref{thm:main_kernel_op} now yields
$F_*\in\Hs_{K_{\mathrm{Szeg\ddot{o}}}}$.
\end{proof}

Our next result applies Theorem~\ref{thm:main_kernel_op} to the parametric elliptic problem
\eqref{eq:elliptic}. The parameters
$c=(c_p)_{p=1}^P$ are chosen according to the properties of the input
coefficient~\eqref{eq:input_coeff}.

\begin{assumption}\label{as:pde_szeg}
Let $(\rho_p)_{p=1}^P\in(0,1)^P$ satisfy
\begin{align}\label{eq:def_delta_admissible}
\sum_{p=1}^P\rho_p^{-1}\abs{\phi_p(s)}
\leq \phi_0(s)-\delta,
\qquad s\in\Omega,
\end{align}
for some $\delta>0$. Moreover, assume that
$\rho_p^2<c_p<1$ for $1\leq p\leq P$.
\end{assumption}

A sequence $(\rho_p)_{p=1}^P$ satisfying
\eqref{eq:def_delta_admissible} is called $\delta$-admissible. Under
Assumption~\ref{as:pde}, such a sequence exists for every
$0<\delta<a_{\mathrm{min}}$; see, for example, \cite[Section~3.1]{cohen2011analytic}.
The following result on the analyticity and Taylor estimates of
$\Fell$ (defined in Section~\ref{sub:parametricPDE}) is crucial for our analysis.

\begin{lemma}[{\cite[Section~2]{cohen2011analytic}}]\label{lem:taylor_ell}\
Under Assumption~\ref{as:pde}, the solution operator $\Fell$ is analytic in
the sense of Definition~\ref{def:analytic_op}. Moreover, under
Assumption~\ref{as:pde_szeg}, its Taylor coefficients satisfy
\begin{align}\label{eq:taylor_est_ell}
\norm{y_{\nu}}_{\Y}
\leq
\frac{C_{\Omega,\Y}\norm{f}_{L^2(\Omega)}}{\delta}\rho^\nu,
\qquad \nu\in\mathbb{N}^P.
\end{align}
Here $C_{\Omega,\Y}=\norm{J}C_\Omega$, where $J\colon H_0^1(\Omega)\to\Y$
is the embedding, $H_0^1(\Omega)$ carries the norm $\norm{\nabla\cdot}_{L^2(\Omega)}$, and $C_\Omega$ is the Poincaré constant.
\end{lemma}

The estimate~\eqref{eq:taylor_est_ell} and
Theorem~\ref{thm:main_kernel_op} immediately yield the following result.

\begin{theorem}[VRKHS for parametric elliptic PDEs]
\label{thm:main_kernel_pde}
Under Assumptions~\ref{as:pde} and~\ref{as:pde_szeg}, the solution operator
$F_*^{\mathrm{ell}}$ belongs to $\Hs_{K_{\mathrm{Szeg\ddot{o}}}}$.
\end{theorem}

\begin{remark}\label{rem:sobolev_emb}
By \cite[Lemma~2.1]{griebel2017reproducing},
$\Hs_{k_{\mathrm{Szeg\ddot{o}}}}\subseteq\Hs_{k_{\mathrm{Sob}}}$ for every Sobolev
order $r>P/2$. Consequently,
$\Hs_{K_{\mathrm{Szeg\ddot{o}}}}\subseteq\Hs_{K_{\mathrm{Sob}}}$, and every
target in $\Hs_{K_{\mathrm{Szeg\ddot{o}}}}$, in particular $\Fell$, also belongs
to the corresponding Sobolev VRKHS.
\end{remark}

For similar results in this direction, see
\cite{griebel2017reproducing, schwab2013qmc, rauhut2017compressive}; in
particular, \cite[Lemma~3.3]{griebel2017reproducing} treats the scalar-valued
case. The present argument extends this reasoning to the vector-valued setting
using the coefficient characterization of separable power-series VRKHSs in
Lemma~\ref{lem:power_series_vrkhs}. The operator setting of
Proposition~\ref{pro:inversion_abstract} also accommodates applications such as
parametric Helmholtz and Maxwell equations.

\begin{remark}\label{rem:assumptions}
The literature on parametric elliptic problems and holomorphic operators
often considers substantially more general settings, including infinitely
many parameters, noncubic parameter domains, nonlinear dependence on the
parameters, and alternative domain or regularity assumptions; see, for
example, \cite{adcock2024learning} and \cite{kuo2017multilevel}. Here we
focus on a setting in which the kernel construction can be explicitly related
to the target operator. Suitable constructions for unbounded or noncubic
domains, and for nonlinear coefficient parametrizations, remain to be
developed. To the best of our knowledge,
\cite{griebel2017reproducing,kempf2019kernel} pursue a related approach under
comparable assumptions, although in less general settings.
\end{remark}

The Gaussian kernel does not yield a VRKHS containing the elliptic solution operators $F_*^{\mathrm{ell}}$ (for $f\neq0$). We provide a simple example here, and discuss the general statement in Appendix~\ref{app:gauss}.
\begin{example}[$F_*^{\mathrm{ell}}\notin \Hs_{K_{\text{Gauss}}}$]\label{ex:gauss}
Take \(P=1\), fix \(0<c<1\), and choose \(\phi_0=1\) and \(\phi_1=-c\), so that \(a(x,s)=1-cx\). These coefficients are Lipschitz continuous and satisfy \(1-c\leq a(x,s)\leq1+c\). For any nonzero \(f\in L^2(\Omega)\), let \(u_0\neq0\) solve \(-\Delta u_0=f\) with homogeneous Dirichlet boundary conditions. The corresponding solution operator is
\(F_*^{\mathrm{ell}}(x)=u_0/(1-cx)\), \(x\in[-1,1]\).
If \(F_*^{\mathrm{ell}}\) belonged to the VRKHS generated by \(K_{\mathrm{Gauss}}=k_{\mathrm{Gauss}}I_{\mathcal Y}\), the tensor-product characterization of separable kernels would imply that
\(x\mapsto\langle F_*^{\mathrm{ell}}(x),u_0\rangle_{\mathcal Y}/\|u_0\|_{\mathcal Y}^2=(1-cx)^{-1}\)
belongs to the scalar Gaussian RKHS. This contradicts the counterexample in Lemma~\ref{lem:gauss_szeg}. Thus, an elliptic solution operator can fail to belong to the Gaussian VRKHS.
\end{example}

\section{Overview of the main results}
\label{sec:overview}

The previous sections introduced the operator-learning setting, the relevant
kernel spaces, and the kernel ridge-regression estimator. We now summarize the
main quantitative conclusions of the error analysis developed in
Section~\ref{sec:proof_details}. The precise assumptions, stronger weighted
metrics, and proofs are given there.

For compactness, let $\bar{F}_{\mathrm{Szeg\ddot{o}}}$ and
$\bar{F}_{\mathrm{Sob}}$ denote the estimators obtained from exact solution
values using the Szegö and Sobolev kernels, respectively. Similarly, let
$\hat{F}_{h,\mathrm{Szeg\ddot{o}}}$ and $\hat{F}_{h,\mathrm{Sob}}$ denote the
corresponding estimators obtained from finite-element solution values with
mesh size $h$.  The following theorem collects the main rates in the
$L^2(\X,\Y)$ metric~\eqref{eq:l2_error}. Technical details are deferred to
Section~\ref{sec:proof_details}, where a complete analysis is given.

\begin{theorem}[Informal main results]
\label{thm:overview_rates}
We  assume the conditions in  Assumptions~\ref{as:sampling} and~\ref{as:bounded_kernel} and  in the noisy setting, in Assumption~\ref{as:noise}.
Here,  $C$ denotes a constant that does not depend on $N$, $h$, or $\eta$. The lower thresholds on $N$ may depend on $\eta$.

\begin{enumerate}[label=(\roman*)]

\item Under the  hypotheses of
Theorem~\ref{thm:main_szeg_operator}, consider the noiseless
setting with $F_*\in\Hs_{K_{\mathrm{Szeg\ddot{o}}}}$. Consider the Szegö kernel and,
for $q>2P$, choose
$\lambda_N^{\mathrm{Szeg\ddot{o}}}=\exp(-N^{1/q})$.
Then, for every
$0<\lambda\leq\lambda_N^{\mathrm{Szeg\ddot{o}}}$ and all sufficiently large $N$,
\begin{equation}
\norm{F_*-\bar{F}_{\mathrm{Szeg\ddot{o}}}}_{L^2(\X,\Y)}
\leq
C\exp\left(-\frac{1}{2}N^{1/q}\right),
\label{eq:overview_szeg_rate}
\end{equation}
with probability at least $1-\eta$.

\item Under the  hypotheses of
Theorem~\ref{thm:main_sob_operator}, consider the noiseless setting with
$F_*\in\Hs_{K_{\mathrm{Szeg\ddot{o}}}}$. Consider a Sobolev kernel
$K_{\mathrm{Sob}}$ of order $r>P/2$, with
$P/(2r)<a\leq1$. Choose
$\lambda_N^{\mathrm{Sob}}=N^{-r'}$ with $0<r'<1/a$.
Then, for every
$0<\lambda\leq\lambda_N^{\mathrm{Sob}}$ and all sufficiently large $N$,
\begin{equation}
\norm{F_*-\bar{F}_{\mathrm{Sob}}}_{L^2(\X,\Y)}
\leq
C N^{-r'/2},
\label{eq:overview_sob_rate}
\end{equation}
with probability at least $1-\eta$.

\item Under the hypotheses of
Theorem~\ref{thm:main_diffusion_rates}, consider the parametric elliptic
solution operator $\Fell$ and suppose that the training outputs are generated
by a finite-element approximation with mesh size $h$, and let $\lambda=\lambda_N$. Then, with probability
at least $1-\eta$,
\begin{equation}
\norm{\Fell-\hat{F}_{h,\mathrm{Szeg\ddot{o}}}}_{L^2(\X,\Y)}
\leq
C\left[
\exp\left(-\frac{1}{2}N^{1/q}\right)
+
h
\right],
\label{eq:overview_pde_szeg_rate}
\end{equation}
for the Szegö kernel, where $q>2P$, and
\begin{equation}
\norm{\Fell-\hat{F}_{h,\mathrm{Sob}}}_{L^2(\X,\Y)}
\leq
C\left[
N^{-r'/2}
+
h
\right]
\label{eq:overview_pde_sob_rate}
\end{equation}
for the Sobolev kernel, where $P/(2r)<a\leq1$ and
$0<r'<1/a$.

\end{enumerate}
\end{theorem}

\begin{remark}[Coupling of mesh size and sample size]
\label{rem:coupling}\
Part~(iii) prescribes how the mesh size has to be coupled to $N$. Choosing
$h\leq\exp\left(-\frac12N^{1/q}\right)=\sqrt{\lambda_N^{\mathrm{Szeg\ddot{o}}}}$,
as in Section~\ref{sec:experimental_overview}, retains the
rate~\eqref{eq:overview_szeg_rate} for exact data, and
$h\leq N^{-r'/2}$ retains the rate~\eqref{eq:overview_sob_rate}.
Thus, the Szegö kernel retains its exponential rate if the mesh is
refined exponentially in $N^{1/q}$, whereas a polynomial refinement suffices
for Sobolev kernels.
\end{remark}

The first two parts of the theorem illustrate how the choice of kernel
affects the learning of analytic operators. For the Szegö kernel, the
polylogarithmic estimate for the maximal effective dimension leads to an
exponential noiseless rate in a  power of $N$
for a suitable choice of the regularization parameter; see
Theorem~\ref{thm:main_szeg_operator}. For Sobolev kernels, the corresponding
polynomial estimate yields an algebraic rate; see
Theorem~\ref{thm:main_sob_operator}. Faster Sobolev rates under additional source conditions are
discussed in Remark~\ref{rem:sobolev_saturation}. Thus, even for
the same analytic target, the learning rate depends on the interaction between
the target, the kernel, and the sampling procedure. The third part quantifies
the additional error arising when the solution operator is accessed through a
numerical solver. In particular, the finite-element error enters
additively and is not amplified by the learning procedure.

The main technical ingredient is the estimation of the maximal effective
dimension for the analytic kernels. The corresponding Szegö and Sobolev
estimates are stated in Propositions~\ref{prop:max_dof_szeg} and
\ref{prop:max_dof_sob}, respectively. The proof of the Szegö estimate, based
on techniques related to those in
\cite{dommel2024bound,dommel2025approximation}, is given in
Appendix~\ref{app:analytic_kernels}. The general error estimate and its
kernel-specific consequences are developed in
Section~\ref{sec:proof_details}; the auxiliary covariance, concentration, and
spectral-factor estimates used there are collected in
Appendix~\ref{app:auxiliary_estimates}. The finite-element contribution is
established in Theorem~\ref{thm:main_diffusion_rates}.

The results are formulated for vector-valued outputs and separable kernels.
Corresponding scalar kernel-ridge-regression results follow as special cases
as described in the remarks after the relevant operator-valued results.
Although the Gaussian kernel is widely used for analytic learning, our
well-specified analysis focuses on the Szegö and Sobolev kernels. The Gaussian
RKHS has a different structure, and $F_*^{\mathrm{ell}}$ does not belong to the Gaussian VRKHS for $f\neq0$, as discussed in
Example~\ref{ex:gauss}. We therefore do not include the Gaussian kernel
among the kernels analyzed in the main results.

Our analysis builds on the classical theory of kernel ridge regression
\cite{caponnetto2007optimal,bauer2007regularization,smale2007learning}
and on recent work on fast rates in the noiseless setting
\cite{cheng2024a,long2024duality,barzilai2024}. We formulate the analysis
directly for vector-valued outputs using separable kernels, drawing on
\cite{li2024towards,meunier2024optimal}. For parametric elliptic problems, our
setting is connected to analytic regularity results
\cite{cohen2010convergence,cohen2011analytic} and to earlier kernel-based
treatments of parametric PDEs
\cite{rieger2017sampling,griebel2017reproducing,kempf2019kernel}. The present
work combines these ingredients to obtain operator-valued learning bounds for
analytic solution operators under random sampling and to quantify the
additional effect of numerical discretization.

\section{Error analysis}
\label{sec:proof_details}

This section provides finite-sample error estimates for operator learning with
kernel ridge regression. We work in the setting of
Section~\ref{subsec:op_learn}, under Assumption~\ref{as:noise} and, when
$\mathcal{X}=[-1,1]^P$, under Assumption~\ref{as:sampling}. We use a separable
kernel $K$ as in~\eqref{sep_ker}, satisfying Assumption~\ref{as:bounded_kernel},
and consider the estimator $\hat{F}$ introduced in~\eqref{vec_krr}, together
with its noiseless counterpart $\bar{F}$. We first establish a general error
estimate based on standard arguments. Its specialization to specific kernels
and regimes, particularly to the learning of analytic operators, constitutes
the main contribution developed later in this section.

\subsection{General metrics and problem invariants}
\label{subsec:operator_quantities}

For $F,G\in\Hs_K$, write
$F(x)=C^*\varphi(x)$ and $G(x)=B^*\varphi(x)$, with
$C,B\in S_2(\Y,\Hs_k)$. Let
$\tpsi\colon[0,\kappa^2]\to\mathbb{R}$ be continuous, and define
\begin{equation}
\label{eq:weighted_operator_norm}
\norm{F-G}_{\tpsi}
=
\norm{\tpsi(\Sigma)(C-B)}_{S_2(\Y,\Hs_k)},
\end{equation}
where functions of $\Sigma$ are defined through  spectral calculus. In general, \eqref{eq:weighted_operator_norm} defines a seminorm; it is a norm if $\tpsi(\Sigma)$ is injective, as for the choices below when $\mu$ has full support. The
choice $\tpsi(\sigma)=\sqrt{\sigma}$ recovers the
$L^2(\X,\Y)$ metric  in~\eqref{eq:l2_error}, see \eqref{vec_norm_eq}. Other
choices of $\tpsi$ tailor the metric to the considered kernels as illustrated by the following remark.

\begin{remark}[Weighted metrics and Sobolev norms]  \label{rem:norms}
Let $\X=\mathbb{T}=\mathbb{R}/2\pi\mathbb{Z}$,
$\mathrm{d}\mu(x)=\mathrm{d}x/(2\pi)$, and  $\Y=\mathbb{R}$. Let
$e_0(x)=1$, $e_{\ell,c}(x)=\sqrt{2}\cos(\ell x)$, and
$e_{\ell,s}(x)=\sqrt{2}\sin(\ell x)$, $\ell\geq1$, be the real Fourier basis.
Writing the corresponding Fourier coefficients of $f$ as
$\widehat f_0$, $\widehat f_{\ell,c}$, and $\widehat f_{\ell,s}$, one has
\[
\norm{f}_{H^t(\mathbb{T})}^2
\asymp
\abs{\widehat f_0}^2+
\sum_{\ell\geq1}(1+\ell^2)^t
\left(
\abs{\widehat f_{\ell,c}}^2+
\abs{\widehat f_{\ell,s}}^2
\right).
\]
Consider first the periodic Sobolev kernel
\[
k_{\mathrm{Sob}}(x,x')
=
1+2\sum_{\ell\geq1}
(1+\ell^2)^{-t}\cos\bigl(\ell(x-x')\bigr),
\qquad t>\frac12.
\]
Its covariance operator has eigenvalues
$\sigma_0^{\mathrm{Sob}}=1$ and
$\sigma_\ell^{\mathrm{Sob}}=(1+\ell^2)^{-t}$ for $\ell\geq1$. The weighted metric has the Fourier weights
$\tpsi(\sigma_\ell^{\mathrm{Sob}})^2/\sigma_\ell^{\mathrm{Sob}}$ (cf.\ the
analytic case below), so that $\tpsi\equiv1$ gives
\(
\norm{f}_{\tpsi}
\asymp
\norm{f}_{H^t(\mathbb{T})},
\)
the Sobolev norm of order $t$. Next, consider the periodic analytic kernel
\[
k_{\mathrm{an}}(x,x')
=
1+2\sum_{\ell\geq1}
\rho^\ell\cos\bigl(\ell(x-x')\bigr),
\qquad 0<\rho<1.
\]
Its covariance operator has eigenvalues
$\sigma_0^{\mathrm{an}}=1$ and
$\sigma_\ell^{\mathrm{an}}=\rho^\ell$ for $\ell\geq1$. The corresponding
$\mathcal{H}_{k_{\mathrm{an}}}$-orthonormal eigenvectors are
$e_0$, $\sqrt{\sigma_\ell^{\mathrm{an}}}\,e_{\ell,c}$, and
$\sqrt{\sigma_\ell^{\mathrm{an}}}\,e_{\ell,s}$. Hence,
\[
\norm{f}_{\tpsi}^2
=
\tpsi(1)^2\abs{\widehat f_0}^2+
\sum_{\ell\geq1}
\frac{\tpsi(\sigma_\ell^{\mathrm{an}})^2}
{\sigma_\ell^{\mathrm{an}}}
\left(
\abs{\widehat f_{\ell,c}}^2+
\abs{\widehat f_{\ell,s}}^2
\right).
\]
Since
$\abs{\log(\sigma_\ell^{\mathrm{an}})}
=\ell\abs{\log(\rho)}$, the choice
$\tpsi(\sigma)=\sqrt{\sigma}(1+\abs{\log(\sigma)})^t$ gives a norm equivalent
to the $H^t(\mathbb{T})$ norm. Thus, the same Sobolev norm can be represented
by different kernels through appropriate spectral functions.
\end{remark}

The regularity of the target operator is described by the following source
condition; see, for example,
\cite{bauer2007regularization,meunier2024optimal}.

\begin{assumption}[Source condition]
\label{as:source_operator}
Let $F_*,F_0\in\Hs_K$ have representations
$F_*(x)=C_*^*\varphi(x)$ and $F_0(x)=C_0^*\varphi(x)$, respectively. We
assume that
\begin{equation}
\label{eq:source_operator}
C_*=\psi(\Sigma)C_0,
\end{equation}
where $\psi\colon[0,\kappa^2]\to\mathbb{R}$ is continuous and non-decreasing.
\end{assumption}

A classical choice is the Hölder-type source condition
$\psi(t)=t^{\beta}$ for some $\beta>0$, which is natural for Sobolev-type
kernels. By contrast, for analytic kernels, more general non-polynomial
source functions can be suitable. The choice $\psi\equiv1$
corresponds to $F_0=F_*$.

The statistical estimates are governed by the following two spectral
quantities.

\begin{definition}[Effective dimensions]
\label{def:effective_dimensions}
For $\lambda>0$, let
$\Sigma_\lambda=\Sigma+\lambda\I_{\Hs_k}$. The effective dimension and the
maximal effective dimension are defined, respectively, by
\begin{equation}
\label{eq:eff_dim_operator}
\mathcal{N}(\lambda)
=
\operatorname{Tr}\bigl(\Sigma_\lambda^{-1}\Sigma\bigr)
=
\sum_{l\geq1}\frac{\sigma_l}{\lambda+\sigma_l},
\end{equation}
and
\begin{equation}
\label{eq:max_eff_dim_operator}
\mathcal{N}_{\infty}(\lambda)
=
\sup_{x\in\X}
\innerpro{\Sigma_\lambda^{-1}\varphi(x),\varphi(x)}_{\Hs_k}.
\end{equation}
\end{definition}

The effective dimension depends on the eigenvalue decay of $\Sigma$, while the
maximal effective dimension also captures the behavior of the corresponding
eigenfunctions. Since $\mathcal{N}(\lambda)=\|\innerpro{\Sigma_\lambda^{-1}\varphi(x),\varphi(x)}_{\Hs_k}\|_{L^1(\mu)}$ and $\mu$ is a probability measure,
$\mathcal{N}(\lambda)\leq\mathcal{N}_{\infty}(\lambda)$.
In what follows, $\norm{\cdot}$ denotes the operator norm on $\Hs_k$ unless
specified otherwise.
With these definitions in place, we formulate the general error estimate that
serves as the starting point for the kernel-specific analyses below.

\subsection{General error estimate}

Learning bounds for the different kernels and regimes considered in this paper
are obtained by specializing the following operator-valued estimate.

\begin{theorem}[General operator error estimate]
\label{thm:general_error}
Assume Assumption~\ref{as:source_operator} and $\Sigma\neq0$, and let
$0<\lambda\leq\lambda_N$. Fix $0<\eta<1$ and define
\begin{equation}
\label{u_def}
u(N,\lambda_N,\eta)
=
\frac{\mathcal{N}_{\infty}(\lambda_N)}{N}
\log\left(
\frac{
4\mathcal{N}_{\infty}(\lambda_N)
\bigl(\lambda_N+\norm{\Sigma}\bigr)
}{
\eta\norm{\Sigma}
}
\right).
\end{equation}
Assume that
\begin{equation}
\label{u_cond}
u(N,\lambda_N,\eta/2)\leq\frac{1}{8}.
\end{equation}
Then, with probability at least $1-\eta$,

\begin{align}
\norm{F_*-\hat{F}}_{\tpsi}
&\leq
\norm{\tpsi(\Sigma)\Sigma_{\lambda_N}^{-1/2}}
\Bigg[
3\lambda_N
\norm{\psi(\Sigma)\Sigma_{\lambda_N}^{-1/2}}
\norm{F_0}_{\Hs_K}
 \nonumber \\ &+
\frac{12L\log(4/\eta)}{\sqrt{N}} \frac{\lambda_N}{\lambda}
\left(
\frac{\sqrt{\mathcal{N}_{\infty}(\lambda_N)}}{\sqrt{N}}  +
\sqrt{\mathcal{N}(\lambda_N)}
\right)
\Bigg].
\label{eq:general_error_bound_operator}
\end{align}
\end{theorem}

Given this result, specific bounds are obtained by choosing the source
function and the weighted norm and by estimating the effective and maximal
effective dimensions for the kernels under consideration. These kernel-specific
estimates, together with their application to analytic operators and
parametric PDEs, constitute the main contribution of our analysis. The proof follows the standard operator argument for kernel ridge regression
developed in~\cite{caponnetto2007optimal} and its separable-kernel formulation
in~\cite{meunier2024optimal}; we formulate it
for general source functions and weighted norms, including non-polynomial
choices. The proof uses known auxiliary estimates collected in
Appendix~\ref{app:auxiliary_estimates}.

\begin{proof}
By Assumption~\ref{as:source_operator},
$C_*=\psi(\Sigma)C_0$. The representer formulas in
Section~\ref{sec:vrkhs_main} give
$\bar{C}=\hat{\Sigma}_{\lambda}^{-1}\hat{\Sigma}C_*$ and
$\hat{C}=\hat{\Sigma}_{\lambda}^{-1}G_z$.
We first decompose the error as
\begin{align}
\norm{F_*-\hat{F}}_{\tpsi}
\leq
\norm{F_*-\bar{F}}_{\tpsi}
+
\norm{\bar{F}-\hat{F}}_{\tpsi}.
\label{eq:est_approx_decomp_operator}
\end{align}
For the first term, using
$C_*-\bar{C}=\lambda\hat{\Sigma}_{\lambda}^{-1}\psi(\Sigma)C_0$
and inserting the regularized covariance operators at the scale $\lambda_N$,
we obtain
\begin{equation}
\norm{F_*-\bar{F}}_{\tpsi}
\leq
\lambda
\norm{\tpsi(\Sigma)\Sigma_{\lambda_N}^{-1/2}}
\norm{\Sigma_{\lambda_N}^{1/2}
\hat{\Sigma}_{\lambda_N}^{-1/2}}^{2}
\norm{\hat{\Sigma}_{\lambda_N}^{1/2}
\hat{\Sigma}_{\lambda}^{-1/2}}^{2}
\norm{\psi(\Sigma)\Sigma_{\lambda_N}^{-1/2}}
\norm{F_0}_{\Hs_K}.
\label{eq:approximation_factorization_operator}
\end{equation}

The factor
\(
\norm{\hat{\Sigma}_{\lambda_N}^{1/2}
\hat{\Sigma}_{\lambda}^{-1/2}}^{2}
\)
is controlled by Lemma~\ref{lem:comp_betw_lambdas}, while
\(
\norm{\Sigma_{\lambda_N}^{1/2}
\hat{\Sigma}_{\lambda_N}^{-1/2}}^{2}
\)
is controlled by Lemma~\ref{lem:bernstein_conc}. Since
$\lambda\leq\lambda_N$, these estimates yield
\[
\norm{F_*-\bar{F}}_{\tpsi}
\leq
3\lambda_N
\norm{\tpsi(\Sigma)\Sigma_{\lambda_N}^{-1/2}}
\norm{\psi(\Sigma)\Sigma_{\lambda_N}^{-1/2}}
\norm{F_0}_{\Hs_K}.
\]
For the second term, the identity
\(
\bar{C}-\hat{C}
=
\hat{\Sigma}_{\lambda}^{-1}
(\hat{\Sigma}C_*-G_z)
\)
and again inserting  $\Sigma_{\lambda_N}$ give
\begin{equation}
\norm{\bar{F}-\hat{F}}_{\tpsi}
\leq
\norm{\tpsi(\Sigma)\Sigma_{\lambda_N}^{-1/2}}
\norm{\Sigma_{\lambda_N}^{1/2}
\hat{\Sigma}_{\lambda_N}^{-1/2}}^{2}
\norm{\hat{\Sigma}_{\lambda_N}^{1/2}
\hat{\Sigma}_{\lambda}^{-1/2}}^{2}
\norm{\Sigma_{\lambda_N}^{-1/2}
(\hat{\Sigma}C_*-G_z)}_{S_2(\Y,\Hs_k)}.
\label{eq:propagation_factorization_operator}
\end{equation}
The concentration estimate in
Lemma~\ref{lem:mom_bound}, applied with confidence parameter $\eta/2$, gives
\[
\norm{\Sigma_{\lambda_N}^{-1/2}
(\hat{\Sigma}C_*-G_z)}_{S_2(\Y,\Hs_k)}
\leq
\frac{4L\log(4/\eta)}{\sqrt{N}}
\left(
\frac{\sqrt{\mathcal{N}_{\infty}(\lambda_N)}}{\sqrt{N}}
+
\sqrt{\mathcal{N}(\lambda_N)}
\right).
\]
Combining the approximation and propagation bounds with
\eqref{eq:est_approx_decomp_operator}, and applying a union bound proves the result.
\end{proof}

Before developing the estimate for different kernel choices, we comment on its
specialization to the scalar case.

\begin{remark}[Scalar case]\
When $\Y=\Real$, the space $S_2(\Real,\Hs_k)$ is canonically identified with
$\Hs_k$. Accordingly, $C$ corresponds to $f\in\Hs_k$, and the representation
$F(x)=C^*\varphi(x)$ reduces to
\(
f(x)=\innerpro{f,\varphi(x)}_{\Hs_k}.
\)
Under this identification, the  metric~\eqref{eq:weighted_operator_norm} becomes
\[
\norm{f-g}_{\tpsi}
=
\norm{\tpsi(\Sigma)(f-g)}_{\Hs_k},
\qquad
f,g\in\Hs_k.
\]
For $\tpsi(\sigma)=\sqrt{\sigma}$, this is the usual
$L^2(\X,\Real)$ norm by~\eqref{scal_norm_eq}. Likewise,
Assumption~\ref{as:source_operator} reduces to the standard scalar source
condition
\(
f_*=\psi(\Sigma)f_0.
\)
The vector-valued kernel ridge-regression estimator~\eqref{vec_krr} becomes the
scalar estimator~\eqref{scal_krr}, and the representer formulas for $\hat{F}$
and $\bar{F}$ reduce to~\eqref{eq:representer_noise} and
\eqref{eq:representer_noiseless}, respectively.  The scalar theory
is recovered as a direct specialization of the operator-valued formulation.
In fact, the proof for separable kernels is essentially the same as the
scalar-case proof, up to these identifications.
\end{remark}

\subsection{Learning analytic operators with the Szegö kernel}
\label{subsec:szeg}

We now specialize the general error estimate to analytic target operators and
the Szegö kernel~\eqref{eq:def_Szeg}. We first establish an estimate for the
maximal effective dimension, which will be used throughout this subsection.
Recall that the Szegö kernel is parametrized by
$c\in (0,1)^P$, and set
\(
c_{\max}=\max_{1\leq p\leq P}c_p.
\)

\begin{proposition}[Maximal effective dimension for the Szegö kernel]
\label{prop:max_dof_szeg}
Consider the Szegö kernel given in
\eqref{eq:def_Szeg}. For sufficiently small $\lambda>0$,
\begin{equation}
\mathcal{N}_{\infty}(\lambda)
\leq
C
\left(\log\frac{1}{\lambda}\right)^{2P},
\label{eq:max_dof_main}
\end{equation}
where $C>0$ depends only on $\mu$, $c_{\max}$, and $P$.
\end{proposition}

The proof, based on approximating the feature map by suitable weight
functions, is given in Appendix~\ref{app:analytic_kernels}. The above
estimate yields an explicit bound for
$u(N,\lambda_N,\eta/2)$ in~\eqref{u_def}. In particular, for
$\lambda_N=\exp(-N^{1/q})$ with $q>2P$, it gives
$u(N,\lambda_N,\eta/2)\leq C N^{\frac{2P}{q}-1}\log(CN/\eta)$ for all sufficiently
large $N$. This tends to zero, so condition~\eqref{u_cond} holds beyond a
threshold that we denote by
$N_{\mathrm{Szeg\ddot{o}}}=N_{\mathrm{Szeg\ddot{o}}}(\eta,q)$; the threshold is understood
to be large enough that the small-$\lambda$ range in
Proposition~\ref{prop:max_dof_szeg} also applies. To measure the error in a metric adapted to the analytic regularity of the
kernel, we make the following choice of spectral function, see Remark \ref{rem:norms}.

\begin{assumption}[Analytic weighted metric]
\label{as:szeg_norm}
For some $t\geq0$, let
\[
\tpsi(\sigma)
=
\sqrt{\sigma}\bigl(1+\abs{\log(\sigma)}\bigr)^t,
\qquad
0<\sigma\leq\kappa^2,
\]
and set $\tpsi(0)=0$. The corresponding weighted metric
is the one defined in~\eqref{eq:weighted_operator_norm}.
\end{assumption}

We can now state the learning bound. We consider the well-specified case
$F_0=F_*$ and $\psi\equiv1$ in the general estimate.

\begin{theorem}[Learning analytic operators with the Szegö kernel]
\label{thm:main_szeg_operator}
Assume Assumptions~\ref{as:sampling}, \ref{as:noise},
\ref{as:bounded_kernel} and \ref{as:szeg_norm}. Let $K=K_{\mathrm{Szeg\ddot{o}}}$ and suppose that
$F_*\in\Hs_{K_{\mathrm{Szeg\ddot{o}}}}$. Let $q>2P$, $t\geq0$, and
$0<\eta<1$. Set
\[
\lambda_N^{\mathrm{Szeg\ddot{o}}}
=
\exp\left(-N^{1/q}\right)
\]
and assume that
$N\geq N_{\mathrm{Szeg\ddot{o}}}(\eta,q)$.
Then, with probability at least $1-\eta$, for
$\lambda=\lambda_N^{\mathrm{Szeg\ddot{o}}}$,
\begin{equation}
\norm{F_*-\hat{F}}_{\tpsi}
\leq
C\left[
N^{t/q}
\exp\left(-\frac{1}{2}N^{1/q}\right)
+
L\log\left(\frac{4}{\eta}\right)
N^{(P+t)/q-1/2}
\right].
\label{eq:main_szeg_operator}
\end{equation}
In the noiseless setting, with probability at least $1-\eta$, for every
$0<\lambda\leq\lambda_N^{\mathrm{Szeg\ddot{o}}}$,
\begin{equation}
\norm{F_*-\bar{F}}_{\tpsi}
\leq
C N^{t/q}
\exp\left(-\frac{1}{2}N^{1/q}\right).
\label{eq:main_szeg_operator_noiseless}
\end{equation}
Here $C>0$ is independent of $N$, $\eta$, and $L$.
\end{theorem}

For fixed $L>0$, the noise term in~\eqref{eq:main_szeg_operator} tends
to zero provided $q>2(P+t)$, whereas the noiseless
bound~\eqref{eq:main_szeg_operator_noiseless} only requires $q>2P$.

\begin{proof}
The maximal effective-dimension estimate in
Proposition~\ref{prop:max_dof_szeg} and the definition of
$N_{\mathrm{Szeg\ddot{o}}}$ ensure that the condition of
Lemma~\ref{lem:bernstein_conc} holds, and hence that the covariance event
required in Theorem~\ref{thm:general_error} has the required probability.
The empirical noise term is controlled by Lemma~\ref{lem:mom_bound}
through the general estimate. In the well-specified
case, and by Lemma~\ref{lem:log_weight} in Appendix~\ref{app:auxiliary_estimates}, since
$\lambda_N^{\mathrm{Szeg\ddot{o}}}\leq e^{-1}$ for $N\geq1$,
\[
\norm{\psi(\Sigma)\Sigma_{\lambda_N^{\mathrm{Szeg\ddot{o}}}}^{-1/2}}
\leq\left(\lambda_N^{\mathrm{Szeg\ddot{o}}}\right)^{-1/2},
\qquad
\norm{\tpsi(\Sigma)\Sigma_{\lambda_N^{\mathrm{Szeg\ddot{o}}}}^{-1/2}}\leq C N^{t/q}.
\]
Together, these estimates show that the approximation term
in Theorem~\ref{thm:general_error} is bounded by
\[
C\left(\lambda_N^{\mathrm{Szeg\ddot{o}}}\right)^{1/2}
\abs{\log\left(\lambda_N^{\mathrm{Szeg\ddot{o}}}\right)}^t
=
C N^{t/q}
\exp\left(-\frac{1}{2}N^{1/q}\right).
\]

Moreover, Proposition~\ref{prop:max_dof_szeg} gives
$\mathcal{N}(\lambda_N^{\mathrm{Szeg\ddot{o}}})\leq\mathcal{N}_{\infty}(\lambda_N^{\mathrm{Szeg\ddot{o}}})\leq CN^{2P/q}$.
Consequently,
\[
\frac{\sqrt{\mathcal{N}_{\infty}
\left(\lambda_N^{\mathrm{Szeg\ddot{o}}}\right)}}{\sqrt{N}}
+
\sqrt{\mathcal{N}
\left(\lambda_N^{\mathrm{Szeg\ddot{o}}}\right)}
\leq
CN^{P/q}.
\]
Using Lemma~\ref{lem:log_weight} once more, the propagation term of
Theorem~\ref{thm:general_error} is bounded by
\[
CL\log(4/\eta)\,N^{(P+t)/q-1/2}.
\]
Combining the approximation and propagation estimates proves
\eqref{eq:main_szeg_operator}. The noiseless estimate
\eqref{eq:main_szeg_operator_noiseless} follows by retaining only the
approximation term.
\end{proof}

\begin{remark}[Dependence on the radius of analyticity]
\label{rem:analyticity_radius}
Suppose that $\norm{y_\nu}_{\Y}\leq M(1+R)^{-\abs{\nu}}$,
$\nu\in\mathbb{N}^P$, for some $M,R>0$; by the Cauchy estimates, this holds
if $F_*$ extends holomorphically to the polydisc
$\{z\in\mathbb{C}^P:\max_p\abs{z_p}<1+R\}$ and is bounded in norm by $M$
there. This is~\eqref{eq:taylor_est} with $\rho_p=(1+R)^{-1}$, and the choice
$c_p=(1+R)^{-1}\in(\rho_p^2,1)$ in the proof of
Theorem~\ref{thm:main_kernel_op} gives
$\norm{F_*}_{\Hs_{K_{\mathrm{Szeg\ddot{o}}}}}^2\leq
M^2\prod_{p=1}^P(1-\rho_p^2/c_p)^{-1}=M^2(1+1/R)^P$. For $t=0$, that is, in the $L^2(\X,\Y)$ metric, since
$\norm{\Sigma^{1/2}\Sigma_\lambda^{-1/2}}\leq1$, the proof of
Theorem~\ref{thm:main_szeg_operator} gives
\eqref{eq:main_szeg_operator_noiseless} with
$C=3\norm{F_*}_{\Hs_{K_{\mathrm{Szeg\ddot{o}}}}}$. Hence, in the noiseless setting
and for $0<\lambda\leq\lambda_N^{\mathrm{Szeg\ddot{o}}}$, with probability at least
$1-\eta$,
\[
\norm{F_*-\bar{F}}_{L^2(\X,\Y)}
\leq
3M\left(1+\frac{1}{R}\right)^{P/2}
\exp\left(-\frac{1}{2}N^{1/q}\right),
\qquad
N\geq N_{\mathrm{Szeg\ddot{o}}}(\eta,q).
\]
Thus the exponent, constrained only by $q>2P$, does not depend on $R$,
which enters through the constant and, via $c_{\max}=(1+R)^{-1}$, through
the threshold $N_{\mathrm{Szeg\ddot{o}}}$.
\end{remark}

\subsection{Learning analytic operators with Sobolev kernels}
\label{subsec:sob}

We next study analytic target operators learned with Sobolev-type kernels. The
relevant spectral properties of these kernels were collected in
Lemma~\ref{lem:sob_char}. We first recall a  bound for the
maximal effective dimension, which is a standard consequence
of item~(iii) of Lemma~\ref{lem:sob_char}; see also
\cite{fischer2020sobolev,zhang2024optimality}. We include the short proof for
completeness.

\begin{proposition}[Maximal effective dimension for Sobolev kernels]
\label{prop:max_dof_sob}
Let $K_{\mathrm{Sob}}$ be the separable kernel induced by a Sobolev-type
kernel of order $r>P/2$. If
$P/(2r)<a\leq1$,
then, for every $\lambda>0$,
\begin{equation}
\mathcal{N}_{\infty}(\lambda)
\leq
C_a\lambda^{-a}.
\label{eq:max_dof_sobolev}
\end{equation}
\end{proposition}

\begin{proof}
By the spectral decomposition of $\Sigma$,
\[
\mathcal{N}_{\infty}(\lambda)
=
\sup_{x\in\X}
\sum_{l\geq1}
\frac{\sigma_l}{\sigma_l+\lambda}e_l(x)^2.
\]
For $0<a\leq1$ and $\sigma,\lambda>0$,
$\sigma/(\sigma+\lambda)\leq(\sigma/\lambda)^a=\lambda^{-a}\sigma^a$.
Hence,
\[
\mathcal{N}_{\infty}(\lambda)
\leq
\lambda^{-a}
\sup_{x\in\X}
\sum_{l\geq1}\sigma_l^a e_l(x)^2
\leq
C_a\lambda^{-a},
\]
where the final inequality is precisely the eigenfunction estimate
\eqref{eq:eigenfcts} in Lemma~\ref{lem:sob_char}.
\end{proof}

For $\lambda_N=N^{-r'}$ with $r'a<1$, the proposition gives
$u(N,\lambda_N,\eta/2)\leq C N^{r'a-1}\log(CN/\eta)$ for all sufficiently
large $N$. This tends to zero, so condition~\eqref{u_cond} holds beyond a
threshold that we denote by
$N_{\mathrm{Sob}}=N_{\mathrm{Sob}}(\eta,r',a)$.

We make the following choice of
spectral function and corresponding metric.

\begin{assumption}[Sobolev weighted metric]
\label{as:sob_norm}
For some $0\leq t\leq1/2$, define
$\tpsi(\sigma)=\sigma^t$ for $\sigma>0$.
For $t>0$, set $\tpsi(0)=0$, while for $t=0$ take
$\tpsi(\sigma)\equiv1$. The corresponding weighted metric is the one defined
in~\eqref{eq:weighted_operator_norm}.
\end{assumption}

We can now state the learning bound. We consider analytic targets
$F_*\in\Hs_{K_{\mathrm{Szeg\ddot{o}}}}$ learned with the Sobolev kernel
$K_{\mathrm{Sob}}$. By Remark~\ref{rem:sobolev_emb}, such targets belong
to $\Hs_{K_{\mathrm{Sob}}}$, that is, they satisfy
Assumption~\ref{as:source_operator} with $\psi\equiv1$ and $F_0=F_*$.

\begin{theorem}[Learning analytic operators with Sobolev kernels]
\label{thm:main_sob_operator}
Assume Assumptions~\ref{as:sampling}, \ref{as:noise},
\ref{as:bounded_kernel} and \ref{as:sob_norm}. Let $K=K_{\mathrm{Sob}}$ and suppose that
$F_*\in\Hs_{K_{\mathrm{Szeg\ddot{o}}}}$. Choose
\[
\lambda_N^{\mathrm{Sob}}=N^{-r'},
\qquad
0<r'<\frac{1}{a},
\qquad
\frac{P}{2r}<a\leq1.
\]
For every $0<\eta<1$, assume that
$N\geq N_{\mathrm{Sob}}(\eta,r',a)$. Then, with probability at least
$1-\eta$,
for $\lambda=\lambda_N^{\mathrm{Sob}}$,
\begin{equation}
\norm{F_*-\hat{F}}_{\tpsi}
\leq
C\left[
N^{-r't}
+
L\log\left(\frac{4}{\eta}\right)
N^{r'(1/2+a/2-t)-1/2}
\right].
\label{eq:main_sob_operator}
\end{equation}
In the noiseless setting, with probability at least $1-\eta$, for every
$0<\lambda\leq\lambda_N^{\mathrm{Sob}}$,
\begin{equation}
\norm{F_*-\bar{F}}_{\tpsi}
\leq
C N^{-r't}.
\label{eq:main_sob_operator_noiseless}
\end{equation}
Here $C>0$ is independent of $N$, $\eta$, and $L$.
\end{theorem}

\begin{proof}
The proof is analogous to that of Theorem~\ref{thm:main_szeg_operator}.
By the definition of $N_{\mathrm{Sob}}$, Proposition~\ref{prop:max_dof_sob}
ensures that the condition in Lemma~\ref{lem:bernstein_conc} holds, so the
covariance event used in Theorem~\ref{thm:general_error} has the required
probability. The empirical noise term is controlled by
Lemma~\ref{lem:mom_bound}.
Since $\psi\equiv1$,
$\norm{\psi(\Sigma)\Sigma_{\lambda_N^{\mathrm{Sob}}}^{-1/2}}\leq(\lambda_N^{\mathrm{Sob}})^{-1/2}$.
By Lemma~\ref{lem:power_weight} in  Appendix~\ref{app:auxiliary_estimates},
$\norm{\tpsi(\Sigma)\Sigma_{\lambda_N^{\mathrm{Sob}}}^{-1/2}}\leq(\lambda_N^{\mathrm{Sob}})^{t-1/2}$.
Consequently, the approximation term in
Theorem~\ref{thm:general_error} is bounded by
\[
C\lambda_N^{\mathrm{Sob}}
\left(\lambda_N^{\mathrm{Sob}}\right)^{-1/2}
\left(\lambda_N^{\mathrm{Sob}}\right)^{t-1/2}
=
C\left(\lambda_N^{\mathrm{Sob}}\right)^{t}
=
C N^{-r't}.
\]
Moreover, Proposition~\ref{prop:max_dof_sob} gives
$\mathcal{N}(\lambda_N^{\mathrm{Sob}})\leq\mathcal{N}_{\infty}(\lambda_N^{\mathrm{Sob}})\leq CN^{r'a}$.
Consequently,
\[
\frac{\sqrt{\mathcal{N}_{\infty}
\left(\lambda_N^{\mathrm{Sob}}\right)}}{\sqrt{N}}
+
\sqrt{\mathcal{N}
\left(\lambda_N^{\mathrm{Sob}}\right)}
\leq
CN^{r'a/2}.
\]
Combining this estimate with
$\norm{\tpsi(\Sigma)\Sigma_{\lambda_N^{\mathrm{Sob}}}^{-1/2}}\leq(\lambda_N^{\mathrm{Sob}})^{t-1/2}=N^{r'(1/2-t)}$
gives the propagation bound
$CL\log(4/\eta)N^{r'(1/2+a/2-t)-1/2}$.
Combining the approximation and propagation estimates proves
\eqref{eq:main_sob_operator}. The noiseless estimate
\eqref{eq:main_sob_operator_noiseless} follows by retaining only the
approximation term.
\end{proof}

For fixed $L>0$, the noise term in~\eqref{eq:main_sob_operator} tends to
zero provided $r'<1/(1+a-2t)$.

We add one remark highlighting a crucial limitation in learning analytic functions, hence operators,
using Sobolev kernels.
\begin{remark}[Source conditions and saturation for Sobolev kernels]
\label{rem:sobolev_saturation}
If $F_*$ additionally satisfies Assumption~\ref{as:source_operator} with
$\psi(\sigma)=\sigma^\beta$ for some $0<\beta\leq1/2$, then
$\norm{\Sigma^\beta\Sigma_\lambda^{-1/2}}\leq\lambda^{\beta-1/2}$ (see the
proof of Lemma~\ref{lem:power_weight}), and the approximation terms
in~\eqref{eq:main_sob_operator}, \eqref{eq:main_sob_operator_noiseless}
and~\eqref{eq:main_diffusion_rates_sob} improve from $N^{-r't}$ to
$N^{-r'(t+\beta)}$; in the $L^2(\X,\Y)$ metric, the rate becomes
$N^{-r'(1/2+\beta)}$ instead of $N^{-r'/2}$.

Larger values of $\beta$ do not lead to faster rates: for $\beta>1/2$, the
factor $\lambda\norm{\Sigma^\beta\Sigma_\lambda^{-1/2}}$ is only of order
$\lambda$. This saturation of kernel ridge regression is not an artifact of
the proof. In the scalar periodic and spherical settings
of~\cite[Theorem~5.2]{long2024duality}, even infinitely smooth targets
satisfying $f_*=\Sigma^sf_0$ for every $s>1/2$ cannot, in general, yield
$L^2$-rates faster than those determined by the polynomial decay of the
eigenvalues; such lower bounds transfer to separable kernels by considering
$F_*(x)=f_*(x)y_0$ with $\norm{y_0}_{\Y}=1$ and noise along $y_0$.

On the other hand, $\beta>0$ is a genuine additional hypothesis. Although
every target in $\Hs_{K_{\mathrm{Szeg\ddot{o}}}}$ lies in every Sobolev space,
$\beta=1/2$ requires it to lie in the range of the integral operator of $k$,
which on a bounded domain imposes boundary conditions that analytic targets
need not satisfy. For this reason, Theorems~\ref{thm:main_sob_operator}
and~\ref{thm:main_diffusion_rates}~(ii) are stated for $\beta=0$.
\end{remark}

\subsection{Learning solution operators for parametric elliptic PDEs}
\label{sec:pde}

We conclude the error analysis by applying the preceding results to the
parametric elliptic problem~\eqref{eq:elliptic}. Under
Assumption~\ref{as:pde}, the solution operator $\Fell$ is well defined and,
by elliptic regularity, satisfies
$\sup_{x\in\X}\|\Fell(x)\|_{H^2(\Omega)}<\infty$.

To generate numerical training data, let
$\Y_h\subset H_0^1(\Omega)$ be a finite-dimensional discretization indexed by
a mesh parameter $h>0$. The Galerkin approximation
$F_{*,h}^{\mathrm{ell}}(x)\in\Y_h$ is defined by
\begin{equation}
\int_\Omega
a(x,s)\nabla F_{*,h}^{\mathrm{ell}}(x)(s)\cdot\nabla v_h(s)\,\mathrm{d}s
=
\int_\Omega f(s)v_h(s)\,\mathrm{d}s,
\qquad
v_h\in\Y_h,\quad x\in\X.
\label{eq:elliptic_discretized}
\end{equation}
Once a basis of $\Y_h$ is fixed, this problem reduces to a finite-dimensional
linear system. For conforming first-order finite elements on a suitable
family of triangulations, standard finite-element estimates give
\begin{equation}
\left\|F_{*,h}^{\mathrm{ell}}(x)-\Fell(x)\right\|_{\Y}
\leq
C_{\mathrm{FEM}}h
\left\|\Fell(x)\right\|_{H^2(\Omega)},
\qquad x\in\X,
\label{eq:fem_bound}
\end{equation}
see, for example, \cite[Theorem~4.6]{quarteroni2009numerical}. Consequently,
there exists a constant $C>0$, independent of $x$ and $h$, such that
\begin{equation}
\left\|F_{*,h}^{\mathrm{ell}}(x)-\Fell(x)\right\|_{\Y}
\leq Ch,
\qquad x\in\X.
\label{eq:fem_uniform_bound}
\end{equation}

We consider training data generated by the numerical solver,
$y_i=F_{*,h}^{\mathrm{ell}}(x_i)$, $i=1,\ldots,N$.
Since the finite-element error is generally deterministic and need not be
centered, it cannot be inserted directly into the centered-noise assumption
\ref{as:noise}. We therefore treat it as a deterministic perturbation of the
noiseless kernel ridge-regression problem for the exact solution operator.

\begin{theorem}[Learning parametric elliptic PDEs from numerical data]
\label{thm:main_diffusion_rates}
Assume Assumptions~\ref{as:sampling}, \ref{as:pde}, \ref{as:pde_szeg}, and \ref{as:bounded_kernel}, and fix $0<\eta<1$. Moreover, suppose that
\eqref{eq:fem_uniform_bound} holds. By Theorem~\ref{thm:main_kernel_pde},
$\Fell\in\Hs_{K_{\mathrm{Szeg\ddot{o}}}}$. Let $\hat F_{h,K}$ denote the
estimator~\eqref{vec_krr} computed from the numerical data with
$\lambda=\lambda_N$, where $\lambda_N$ is specified below for each kernel.

\begin{enumerate}[label=(\roman*)]

\item Let $K=K_{\mathrm{Szeg\ddot{o}}}$. Let
$t\geq0$, $q>2P$, and choose
$\lambda_N^{\mathrm{Szeg\ddot{o}}}=\exp(-N^{1/q})$.
If $N\geq N_{\mathrm{Szeg\ddot{o}}}(\eta,q)$,
then, with probability at least $1-\eta$,
\begin{equation}
\left\|\Fell-\hat{F}_{h,\mathrm{Szeg\ddot{o}}}\right\|_{\tpsi}
\leq
C\left[
N^{t/q}\exp\left(-\frac12N^{1/q}\right)
+
hN^{t/q}
\right],
\label{eq:main_diffusion_rates_szeg}
\end{equation}
where $\tpsi$ is the spectral function from
Assumption~\ref{as:szeg_norm}.

\item Let $K=K_{\mathrm{Sob}}$. Let \(r>P/2\) be the Sobolev order,
let \(P/(2r)<a\leq1\), and choose
$\lambda_N^{\mathrm{Sob}}=N^{-r'}$ with $0<r'<1/a$.
If $N\geq N_{\mathrm{Sob}}(\eta,r',a)$, then, with probability at least
$1-\eta$,
\begin{equation}
\left\|\Fell-\hat{F}_{h,\mathrm{Sob}}\right\|_{\tpsi}
\leq
C\left[
N^{-r't}
+
hN^{r'(1/2-t)}
\right],
\label{eq:main_diffusion_rates_sob}
\end{equation}
where $\tpsi(\sigma)=\sigma^t$, \(0\leq t\leq1/2\), is the spectral
function from Assumption~\ref{as:sob_norm}.

\end{enumerate}

Here $C>0$ is independent of $N$, $h$, and $\eta$. In particular, \(t=0\) in
\eqref{eq:main_diffusion_rates_szeg} and \(t=1/2\) in
\eqref{eq:main_diffusion_rates_sob} correspond to the
$L^2(\X,\Y)$ error; in particular, the finite-element error enters
additively.

\end{theorem}

\begin{proof}
We give the argument for a general separable kernel. Let $C_*$ be the
coefficient operator of $\Fell$, and define
\[
\Delta_i
=
F_{*,h}^{\mathrm{ell}}(x_i)-\Fell(x_i),
\qquad
D_h
=
\frac1N\sum_{i=1}^N
\varphi(x_i)\otimes\Delta_i.
\]
The coefficient operator associated with the estimator obtained from the
numerical data is
$\hat{C}_{h,K}=\hat{\Sigma}_{\lambda}^{-1}(\hat{\Sigma}C_*+D_h)$,
whereas the coefficient operator of the noiseless estimator based on the
exact solution values is
$\bar{C}_K=\hat{\Sigma}_{\lambda}^{-1}\hat{\Sigma}C_*$.
Consequently,
$\hat{C}_{h,K}-\bar{C}_K=\hat{\Sigma}_{\lambda}^{-1}D_h$.

Set $\lambda=\lambda_N$. By the definitions of the thresholds in the two
parts of the theorem, the covariance event of
Lemma~\ref{lem:bernstein_conc} has probability at least
$1-\eta$. On this event,
$\|\Sigma_{\lambda_N}^{1/2}\hat{\Sigma}_{\lambda_N}^{-1/2}\|\leq \sqrt3$.
To estimate the perturbation, define $S\colon\mathbb{R}^N\to\Hs_k$ and
$U\colon\Y\to\mathbb{R}^N$ by
$Sv=N^{-1/2}\sum_{i=1}^Nv_i\varphi(x_i)$ and
$(Uy)_i=N^{-1/2}\innerpro{\Delta_i,y}_{\Y}$. Then $D_h=SU$, $SS^*=\hat\Sigma$,
hence $\norm{\hat\Sigma_{\lambda_N}^{-1/2}S}\leq1$, and
$\norm{U}_{S_2(\Y,\mathbb{R}^N)}^2=N^{-1}\sum_{i=1}^N\norm{\Delta_i}_{\Y}^2\leq C^2h^2$
by~\eqref{eq:fem_uniform_bound}. Therefore,
\begin{equation}
\left\|\hat{F}_{h,K}-\bar{F}_K\right\|_{\tpsi}
=
\left\|\tpsi(\Sigma)\hat\Sigma_{\lambda_N}^{-1}SU\right\|_{S_2(\Y,\Hs_k)}
\leq
\sqrt3\,Ch
\left\|\tpsi(\Sigma)
\Sigma_{\lambda_N}^{-1/2}\right\|.
\label{eq:fem_perturbation}
\end{equation}

For the Szegö kernel, Lemma~\ref{lem:log_weight} gives
$\|\tpsi(\Sigma)\Sigma_{\lambda_N^{\mathrm{Szeg\ddot{o}}}}^{-1/2}\|\leq CN^{t/q}$.
The approximation part of Theorem~\ref{thm:main_szeg_operator}, applied to the
exact noiseless data on the same covariance event, gives
$\|\Fell-\bar{F}_{\mathrm{Szeg\ddot{o}}}\|_{\tpsi}\leq CN^{t/q}\exp(-\frac12N^{1/q})$.
Combining this estimate with~\eqref{eq:fem_perturbation} proves~\eqref{eq:main_diffusion_rates_szeg}.

For the Sobolev kernel, Lemma~\ref{lem:power_weight} gives
$\|\tpsi(\Sigma)\Sigma_{\lambda_N^{\mathrm{Sob}}}^{-1/2}\|\leq N^{r'(1/2-t)}$.
The approximation part of Theorem~\ref{thm:main_sob_operator}, applied to the
exact noiseless data on the same covariance event, gives
$\|\Fell-\bar{F}_{\mathrm{Sob}}\|_{\tpsi}\leq CN^{-r't}$.
Combining this estimate with~\eqref{eq:fem_perturbation} proves~\eqref{eq:main_diffusion_rates_sob}.
\end{proof}

\section{Experiments}\label{sec:experimental_overview}

We investigate the influence of kernel choice when learning analytic targets in two settings: noiseless scalar kernel ridge regression (KRR) and vector-valued KRR for a parametric elliptic problem. In all experiments, hyperparameters are selected by cross-validation, and each experiment is repeated ten times on independently drawn data; all computations are carried out in double precision (the finite-element solutions are stored in single precision). The notebooks reproducing all experiments of this section, together with the raw results underlying Figures~\ref{fig:scalar_KRR}--\ref{fig:PDE_var_P} and Table~\ref{tab:results}, are available in the accompanying git repository, \url{https://github.com/markush314/Kernel-based-learning-of-analytic-operators-and-parametric-PDEs}.

\subsection{Scalar KRR}\label{sec:experimental_overview_scalar}

\begin{figure}[htbp]
\centering
\begin{subfigure}[b]{0.44\textwidth}
\centering
\maybeincludegraphics[width=\textwidth]{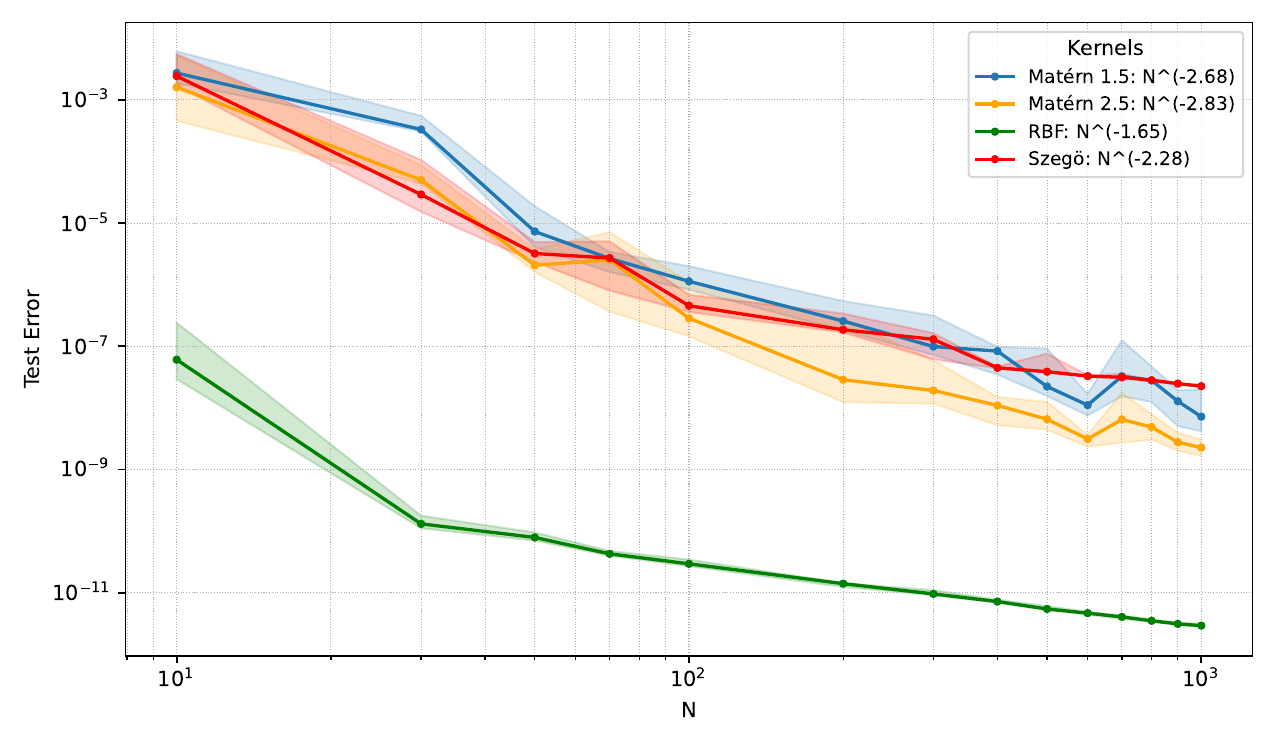}
\caption{$P=1$, $f_{*,1}(x)=e^{-\norm{x}^2}$}
\end{subfigure}
\hspace{0.05\textwidth}
\begin{subfigure}[b]{0.44\textwidth}
\centering
\maybeincludegraphics[width=\textwidth]{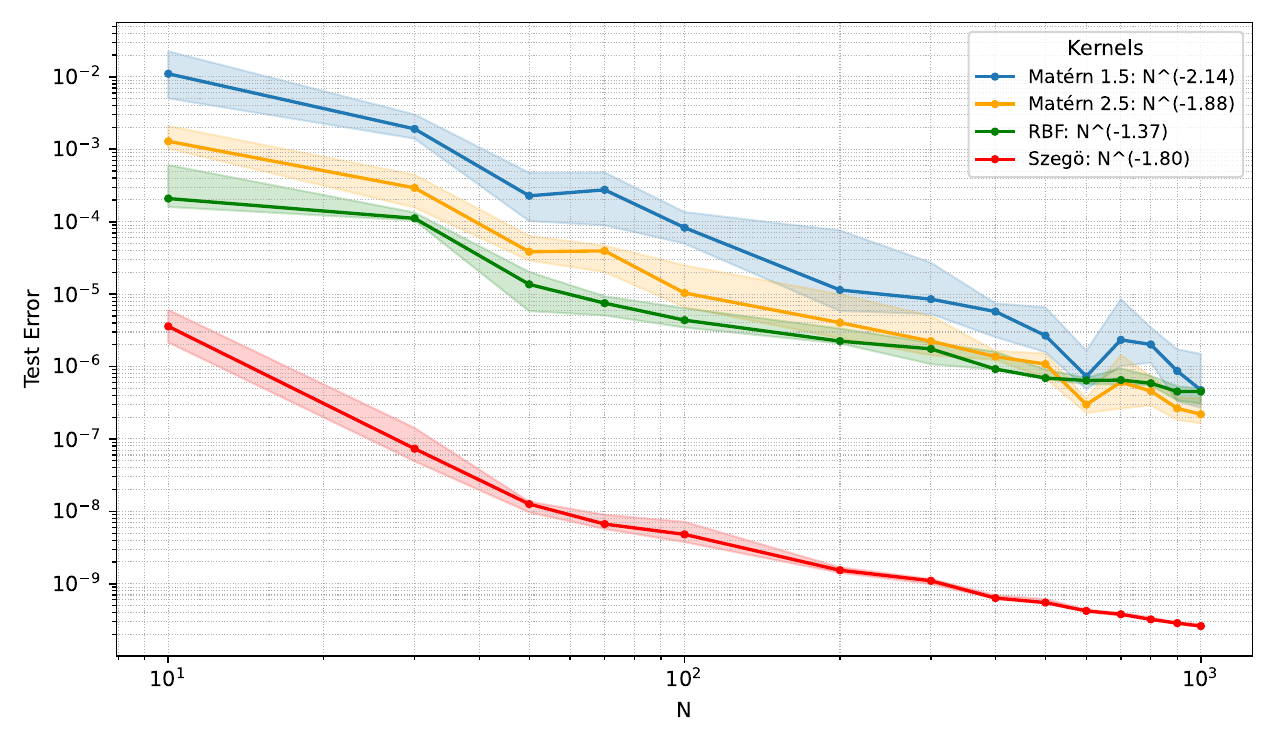}
\caption{$P=1$, $f_{*,2}(x)=\norm{x}^2$}
\end{subfigure}

\vspace{0.3cm}

\begin{subfigure}[b]{0.44\textwidth}
    \centering
    \maybeincludegraphics[width=\textwidth]{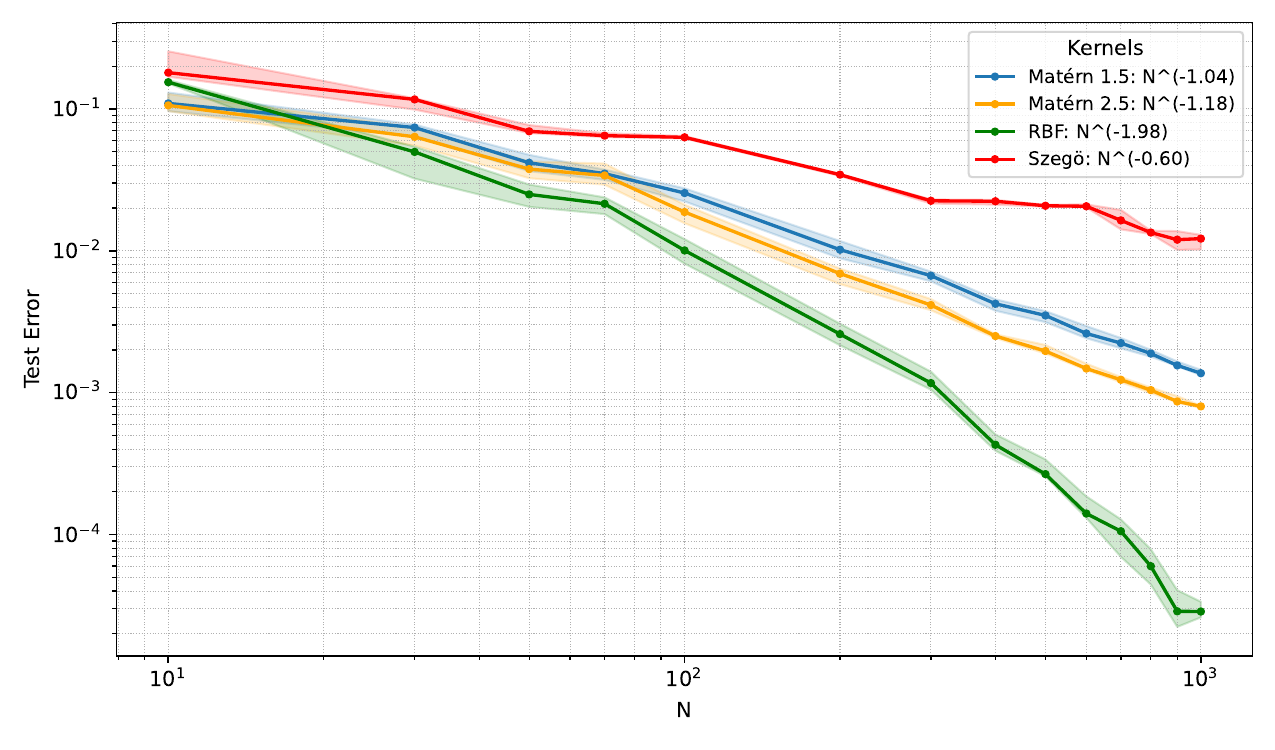}
    \caption{$P=5$, $f_{*,1}(x)=e^{-\norm{x}^2}$}
\end{subfigure}
\hspace{0.05\textwidth}
\begin{subfigure}[b]{0.44\textwidth}
    \centering
    \maybeincludegraphics[width=\textwidth]{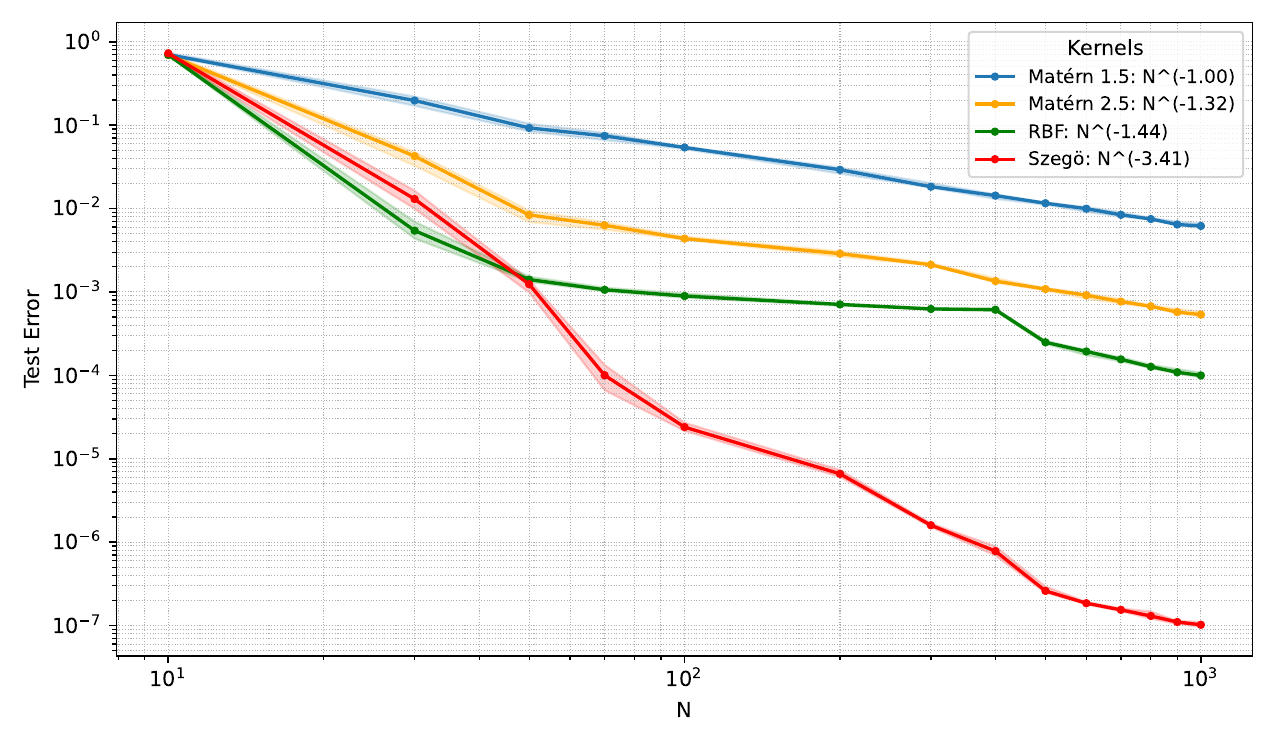}
    \caption{$P=5$, $f_{*,2}(x)=\norm{x}^2$}
\end{subfigure}

\caption{Test error for noiseless scalar KRR in different scenarios: median over ten repetitions, with a tube spanning the interquartile range.}
\label{fig:scalar_KRR}
\end{figure}

We consider $\mathcal{X}=[-1,1]^P$, with $P\in\{1,5\}$, equipped with the uniform measure $\mu$, and $\mathcal{Y}=\mathbb{R}$. We study the analytic target functions $f_{*,1}(x)=e^{-\norm{x}^2}$ and $f_{*,2}(x)=\norm{x}^2$. It is known that $f_{*,1}\in\Hs_{k_{\mathrm{Gauss}}}$ for every scaling $\bar s>1/\sqrt{2}$, i.e., for $\bar s\geq1$ in the grid used below, and that $f_{*,2}\in\Hs_{k_{\mathrm{Szeg\ddot{o}}}}$, whereas $f_{*,2}\notin\Hs_{k_{\mathrm{Gauss}}}$ for every $\bar s$, since polynomials are not contained in Gaussian RKHSs; see, for example, \cite[Theorem 2]{minh2010some}. We compare Gaussian, Matérn kernels with smoothness parameters $\bar r\in\{\frac32,\frac52\}$, and Szegö kernels. Throughout both experiments of this section we use one common pair of hyperparameter grids,
\begin{equation}
\begin{aligned}
\bar s&\in\{0.003,0.01,0.03,0.1,0.3,1,3,10,30,100\},\\
c&\in\{0.01,0.02,0.05,0.1,0.3,0.5,0.7,0.9,0.99\},
\end{aligned}
\label{eq:shared_grid}
\end{equation}
a logarithmic grid spanning about four and a half decades for the scaling parameter $\bar s$ of~\eqref{eq:gauss} and~\eqref{eq:matern}, and a grid covering the admissible range for the Szegö parameter. The Szegö kernel is used in the product form~\eqref{eq:def_Szeg} with a common parameter $c_p=c$ for all $p=1,\ldots,P$. The regularization parameter, added directly to the kernel matrix  and hence equal to $N\lambda$ in the normalization of~\eqref{scal_krr}, is selected from $\{10^{-3},10^{-5},10^{-7},10^{-9}\}$ by three-fold cross-validation; consistent with the noiseless setting, the smallest value is selected in about $94\%$ of all fits. For $N\geq700$ the most frequently selected value is interior to~\eqref{eq:shared_grid} for every kernel and every panel. The test error is computed on $N_{\mathrm{Test}}=20000$ independent samples, for training-set sizes $10\leq N\leq1000$. Figure~\ref{fig:scalar_KRR} reports the median test error over the ten repetitions, together with a tube spanning the interquartile range; observed convergence rates are obtained by log-linear regression of the medians, following \cite{zhang2024optimality}; the test error is the root-mean-square error.
The results show rapid convergence and a clear advantage for kernels adapted to the target: the Gaussian kernel performs best for $f_{*,1}$, while the Szegö kernel performs best for $f_{*,2}$. In both cases, the error decreases by several orders of magnitude, to between about $10^{-12}$ ($P=1$) and $10^{-5}$ ($P=5$) at $N=1000$. All experiments are implemented in Python using scikit-learn \cite{scikit-learn}.

\subsection{Parametric elliptic problem}

We next consider a specific instance of the parametric elliptic problem introduced in Section~\ref{sec:pde}, following \cite[Appendix B.1]{adcock2024optimal} with slight modifications. We take $\Omega=[0,1]^2$, $\mathcal{X}=[-1,1]^P$, $\mathcal{Y}=L^2(\Omega)$, and

$$
a(x,s)=1.89+\sum_{p=1}^P x_p
\frac{\sin(\pi p s_1)}{p^{9/5}},
\qquad
f(s)=10.
$$

For $P=10$, the sequence $\rho_p=0.999\,p^{-1/5}$, $p=1,\ldots,10$, is $\delta$-admissible, since
\begin{equation}\label{eq:p10}
\sum_{p=1}^{10}\rho_p^{-1}|\phi_p(s)|
\leq
0.999^{-1}\sum_{p=1}^{10}p^{1/5-9/5}
<1.8813<1.89.
\end{equation}
Hence, by Theorem~\ref{thm:main_kernel_pde}, any choice with $0.999^2p^{-2/5}<c_p<1$ is covered by the theory. Throughout, we use the problem-adapted family $c_p=c_{\mathrm{scale}}\,0.999^2p^{-2/5}$, in which the anisotropic profile $p^{-2/5}$ is dictated by the coefficient and $c_{\mathrm{scale}}>0$ acts as a single scalar tuning parameter. Replacing $\rho_p$ by $(1-\varepsilon)\sqrt{c_{\mathrm{scale}}}\,0.999\,p^{-1/5}$, $\varepsilon>0$ small, shows that this family is covered by Theorem~\ref{thm:main_kernel_pde} whenever $c_{\mathrm{scale}}>(1.8813/1.89)^2\approx0.991$; the sharper bound~\eqref{eq:sharp_admissible} below relaxes this to $c_{\mathrm{scale}}>0.928$. Smaller values are also included in the grids to probe the sensitivity of the results to this parameter.

\begin{figure}[tbp]
\centering
\maybeincludegraphics[width=0.7\linewidth]{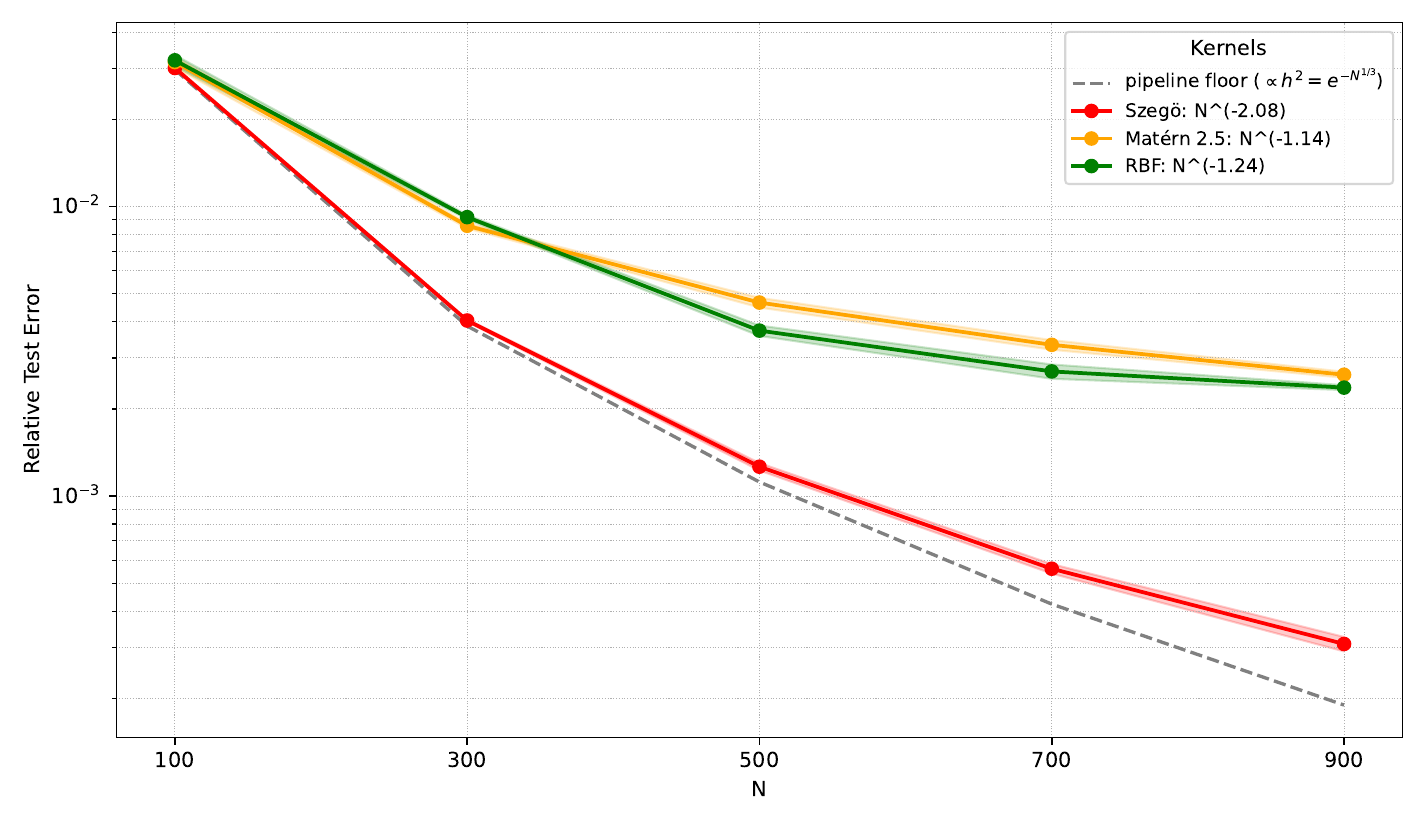}
\caption{Relative test error for $P=10$ with varying $N$ and adapted training resolution, together with the accuracy floor of the data-generation pipeline (dashed). The Szegö kernel stays close to the floor, while the empirical rates of the other kernels are estimation-limited for larger $N$. The slopes in the legend are least-squares fits over the displayed range of $N$.}
\label{fig:PDE_P=10}
\end{figure}

\paragraph{Data generation}
The training pairs $\bigl(x_i,F_{*,h_{\mathrm{Train}}}^{\mathrm{ell}}(x_i)\bigr)_{i=1}^N$ are generated using a uniformly triangulated mesh, first-order Lagrange finite elements, and mesh size $h_{\mathrm{Train}}=1/\lfloor e^{0.5N^{1/3}}\rfloor\approx{}e^{-0.5N^{1/3}}$, so that the discretization is coupled to the sample size as $h=\sqrt{\lambda_N}$, in the spirit of Theorem~\ref{thm:main_diffusion_rates}.
Here $q=3$ is below the value $q>2P$ required by the theorem, for which
the mesh size would hardly vary over the considered $N$; the experiment
thus illustrates this schedule rather than the asymptotic rate. The coefficients $\phi_p$ are interpolated piecewise linearly on each mesh. This is an $O(h^2)$ effect for $p<1/h_{\mathrm{Train}}$, whereas modes with $p\geq1/h_{\mathrm{Train}}$ are aliased; both effects are contained in the pipeline floor below, which is computed with the same solver.
We consider $N\in\{100,300,500,700,900\}$ and draw, for each $N$, ten independent training pools, one per repetition. %
The test set consists of $N_{\mathrm{Test}}=1000$ parameters, whose solutions are computed on a fixed reference mesh with $h_{\mathrm{Test}}=1/320$, using a conjugate-gradient solver with algebraic multigrid preconditioning at relative tolerance $10^{-12}$. The reference mesh is finer than all training meshes: even for the finest training mesh ($N=900$, $h_{\mathrm{Train}}\approx1/124$) the squared mesh-size ratio is $(h_{\mathrm{Test}}/h_{\mathrm{Train}})^2\approx0.15$, so that the reference error is smaller than, but not negligible compared with, the training discretization error at the largest $N$. In addition, the test parameters are solved on each training mesh; comparing these solutions with the reference yields the \emph{accuracy floor} of the data-generation pipeline, that is, the error that would be incurred if the training-mesh solutions were reproduced exactly. Although the theory suggests a polynomially decaying mesh size for the Matérn kernel, we use the exponential schedule for all kernels, as it performs better empirically and allows for a direct comparison. The implementation uses FEniCS \cite{alnaes2015fenics}. Unlike \cite{adcock2024optimal}, which uses mixed finite elements \cite{gatica2014simple}, we use the standard finite-element discretization described in Section~\ref{sec:pde}. We also adapt the implementation of \cite{batlle2024kernel}, including an explicit implementation of the Szegö kernel and an extension of the JAX implementation to accommodate the high-dimensional, fine-resolution setting.

\paragraph{Evaluation}
We measure performance using the relative test error
$$
\frac{1}{N_{\mathrm{Test}}}
\sum_{j=1}^{N_{\mathrm{Test}}}
\frac{
\norm{
F^{\mathrm{ell}}_{*,h_{\mathrm{Test}}}(x_j)-\widehat F(x_j)
}_{L^2(\Omega)}
}{
\norm{
F^{\mathrm{ell}}_{*,h_{\mathrm{Test}}}(x_j)
}_{L^2(\Omega)}
},
$$
where the $L^2(\Omega)$ norm is approximated using the trapezoidal rule. Since
$\norm{\Fell(x)}_{L^2(\Omega)}$ is bounded from above and below uniformly in
$x\in\X$, the population version of this relative error is bounded, up to a constant and by Jensen's inequality, by the absolute
$L^2(\X,\Y)$ error considered in the theory, up to the errors of the reference solution, of the quadrature, and of the transfer described next; the test average is a Monte Carlo estimate of it. Since the learned output $\widehat F(x_j)$ is a piecewise linear finite-element function on the training mesh, it is transferred to the nodes of the reference mesh by bilinear interpolation. For finite-element solutions of sufficiently regular problems, this transfer is of order $h^2$ and therefore matches the accuracy of the finite-element solutions themselves in the norm of $\Y=L^2(\Omega)$, for which the estimate \eqref{eq:fem_bound} improves to order $h^2$ by a duality argument.

\begin{table}[tbp]
\centering
\begin{tabular}{c|ccc|c}
$N$ & Szegö & Matérn & Gaussian & pipeline floor \\
\hline
100 & \textbf{0.03013 $\pm$ 0.00034} & 0.0316 $\pm$ 0.0013 & 0.0320 $\pm$ 0.0014 & 0.029750 \\
300 & \textbf{0.004041 $\pm$ 0.000042} & 0.00859 $\pm$ 0.00017 & 0.00920 $\pm$ 0.00016 & 0.003867 \\
500 & \textbf{0.001263 $\pm$ 0.000038} & 0.00466 $\pm$ 0.00018 & 0.00373 $\pm$ 0.00015 & 0.001118 \\
700 & \textbf{0.000561 $\pm$ 0.000021} & 0.00333 $\pm$ 0.00013 & 0.00270 $\pm$ 0.00015 & 0.000423 \\
900 & \textbf{0.000308 $\pm$ 0.000018} & 0.002625 $\pm$ 0.000064 & 0.002369 $\pm$ 0.000047 & 0.000190 \\
\end{tabular}
\caption{Relative test errors for different kernels as a function of the training-set size $N$ for $P=10$, together with the accuracy floor of the data-generation pipeline. Standard deviations are taken over ten repetitions on independent training pools, and bold entries indicate significantly lower errors according to paired $t$-tests at level $0.01$.
}
\label{tab:results}
\end{table}

\begin{figure}[tbp]
\centering
\begin{minipage}[t]{0.485\linewidth}
\centering
\maybeincludegraphics[width=\linewidth]{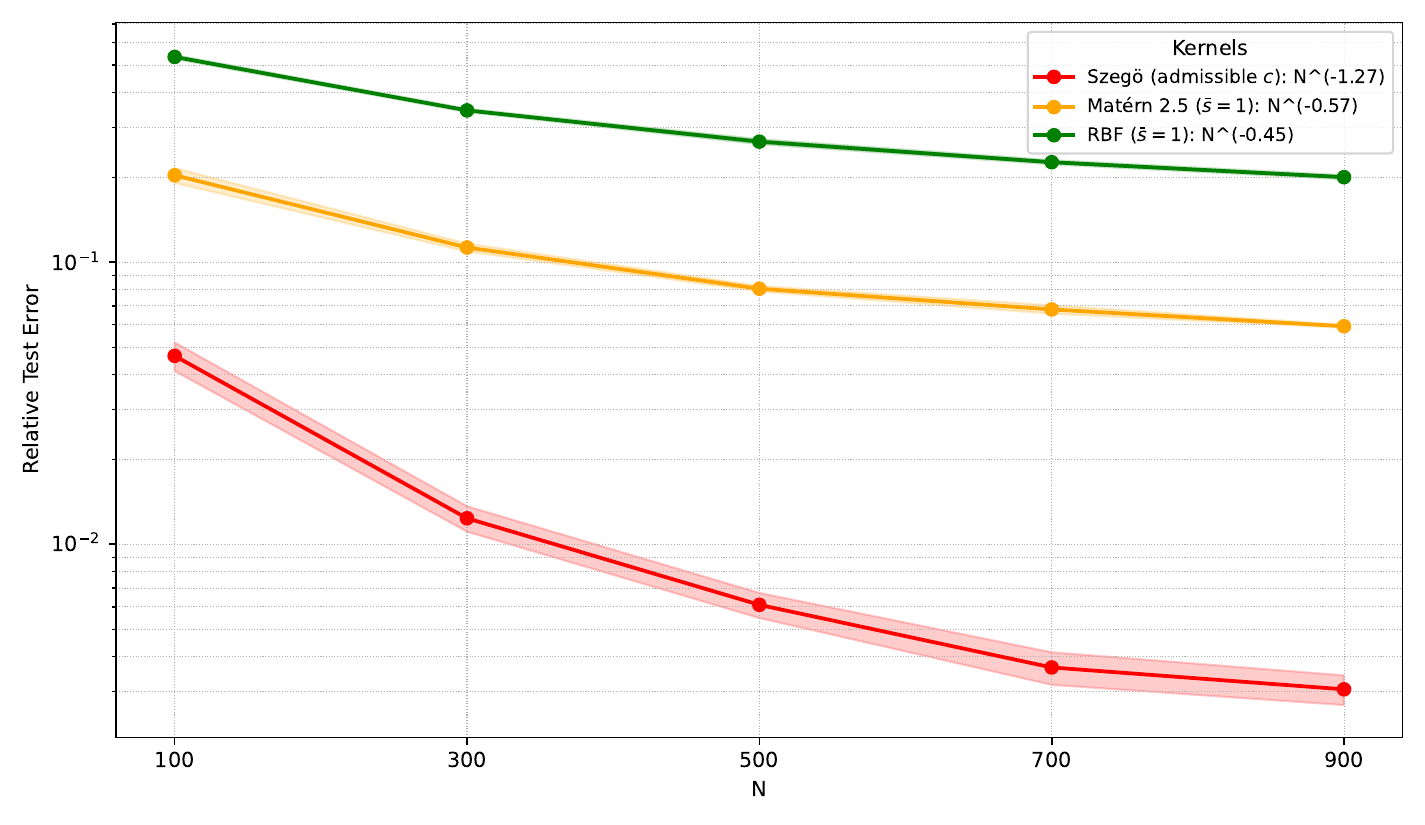}
\caption{Relative test error for $P=10$ with varying $N$ and adapted training resolution. The Szegö kernel with an admissible parameter is compared with the Matérn and Gaussian kernels using the poorly scaled parameter $\bar s=1$.}
\label{fig:PDE_P=10_bad_params}
\end{minipage}\hfill
\begin{minipage}[t]{0.485\linewidth}
\centering
\maybeincludegraphics[width=\linewidth]{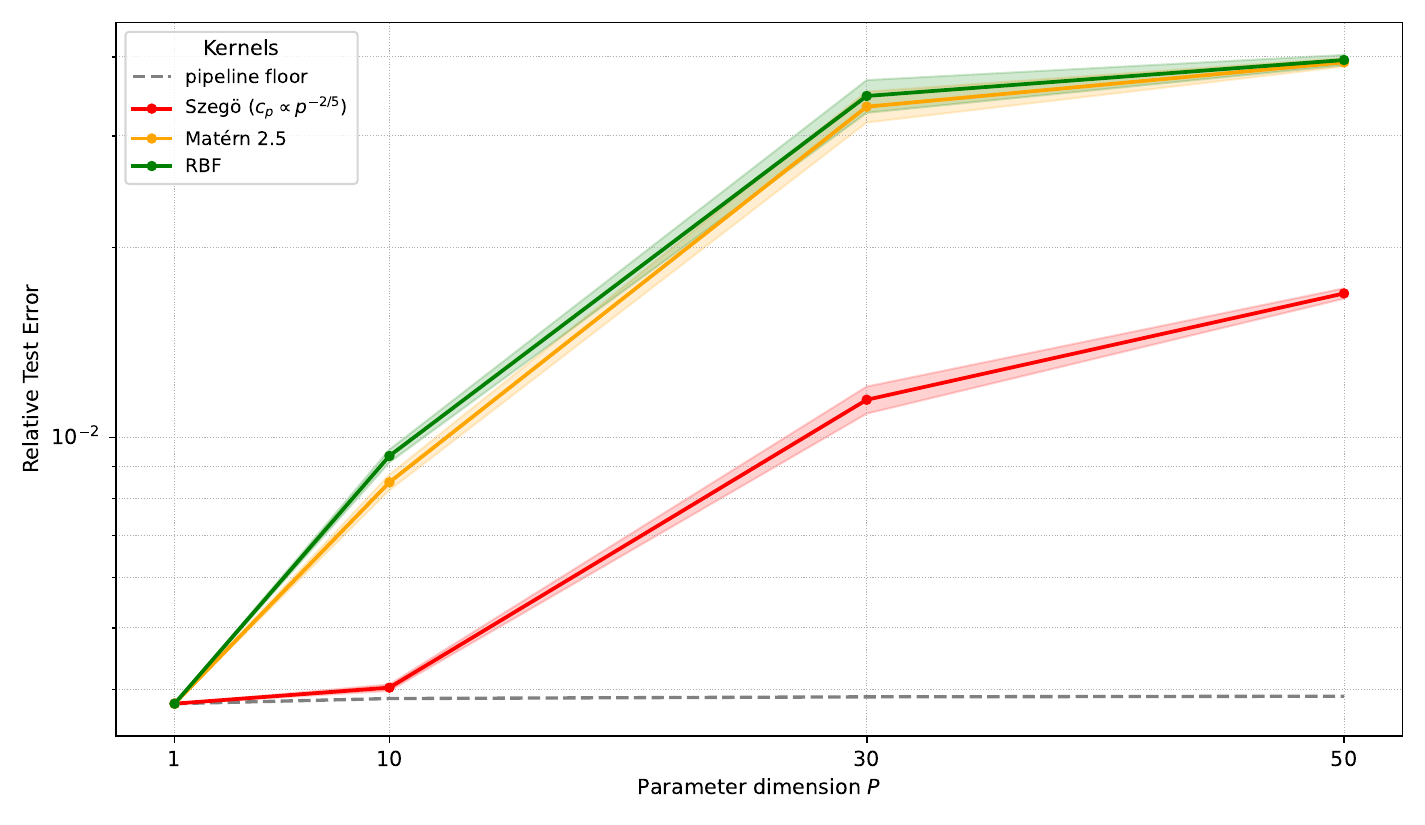}
\caption{Relative test error for varying input dimension $P$ with fixed training-set size $N=300$, together with the pipeline floor (dashed), which is essentially constant in $P$. At $P=1$ all kernels attain the floor, so this data point carries no comparative information.}
\label{fig:PDE_var_P}
\end{minipage}
\end{figure}

\paragraph{Training.}
We learn the operator $F_*^{\mathrm{ell}}$ by vector-valued KRR as in \eqref{vec_krr}. For the Gaussian and the Matérn $\tfrac52$ kernel, and for the Szegö kernel, we select the shared grids~\eqref{eq:shared_grid}, with $c_{\mathrm{scale}}$ in place of $c$; of these the largest value $c_{\mathrm{scale}}=0.99$ is admissible in the sense discussed above. The cross-validated choice is interior in every case: $\bar s\in\{0.01,0.03,0.1\}$ for Matérn and Gauss kernels and $c_{\mathrm{scale}}\in\{0.3,0.5\}$ for the Szegö kernel; the latter values lie below the range certified by Theorem~\ref{thm:main_kernel_pde}, while the certified value $c_{\mathrm{scale}}=0.99$ is used in Figure~\ref{fig:PDE_P=10_bad_params}. The regularization parameter, added directly to the kernel matrix and hence equal to $N\lambda$ in the normalization of~\eqref{vec_krr}, is selected from $\{10^{-3},10^{-5},10^{-7},10^{-9}\}$ by five-fold cross-validation on the mean absolute $L^2(\Omega)$ error, and the Szegö kernel always selects the smallest value, Matérn and Gauss kernels mostly so (otherwise $10^{-7}$ or $10^{-5}$); in all cases $\lambda$ is far below $\lambda_N$. Re-tuning every kernel against the test error on a much denser grid (three repetitions) leaves the Szegö kernel ahead by a factor of two or more at $N=300$ and $N=900$, so the comparison does not hinge on the choice among the tested grids.

\paragraph{Results}
The results for $P=10$ are reported in Figure~\ref{fig:PDE_P=10} and Table~\ref{tab:results}, where significance is assessed by paired $t$-tests at level $0.01$ per comparison. The Szegö kernel stays within $5\%$ of the pipeline floor for $N\leq300$ and within a factor of two for all $N$: its error largely follows the imposed resolution schedule $h^2=e^{-N^{1/3}}$, qualitatively consistent with the fast noiseless rates predicted by the theory. The Matérn and Gaussian kernels, in contrast, remain up to fourteen times as large as the floor despite tuning. A further practical advantage of the Szegö kernel is that a theoretically admissible choice of its parameter performs well without extensive tuning, whereas the Gaussian and Matérn kernels do not have comparably explicit parameter-selection rules. This is illustrated in Figure~\ref{fig:PDE_P=10_bad_params}, where the admissible Szegö parameter is compared with the poorly scaled choice $\bar s=1$ for the other two kernels, only the regularization parameter being cross-validated.

\paragraph{Increasing dimensionality}
To examine the effect of the input dimension, we fix $N=300$ and consider $P\in\{1,10,30,50\}$, with $N_{\mathrm{Test}}=1000$ test samples for each $P$ and all remaining choices, including the grids~\eqref{eq:shared_grid}, as above. For the Szegö kernel we retain the problem-adapted family $c_p=c_{\mathrm{scale}}\,0.999^2p^{-2/5}$, whose admissible range is covered by Theorem~\ref{thm:main_kernel_pde} uniformly in $P$: while the crude estimate $|\sin(\pi ps_1)|\leq1$ verifies the $\delta$-admissibility of $\rho_p=0.999\,p^{-1/5}$ only up to $P=10$ (see \eqref{eq:p10}), the sharp pointwise evaluation of \eqref{eq:def_delta_admissible} yields
\begin{align}
\sup_{s\in\Omega}\sum_{p=1}^{P}\rho_p^{-1}|\phi_p(s)|
&\leq
\sup_{t\in[0,1]}\frac{1}{0.999}\sum_{p=1}^{50}p^{-9/5+1/5}|\sin(\pi pt)|
+\frac{1}{0.999}\cdot\frac53\cdot50^{-3/5}\notag\\
&\leq
1.6606+0.1596<1.89-0.069
\label{eq:sharp_admissible}
\end{align}
for every finite $P$, with $\delta=0.069$. The first term on the right-hand
side bounds the contribution of $p\leq50$; its supremum over $t$ is computed
numerically on a grid, with the discretization error controlled by the
Lipschitz constant of the sum. The second term bounds the tail $p>50$, using
$|\sin|\leq1$ and comparison with an integral. The results are shown in Figure~\ref{fig:PDE_var_P}. The adapted Szegö family remains closest to the floor at every $P>1$, outperforming the other kernels by a factor between $2.1$ and $3.0$.

\paragraph{Further baselines}
Appendix~\ref{app:further_experiments} compares the kernels of this section with
coordinate-weighted (anisotropic) Gaussian and Matérn kernels, with a Szegö kernel whose
anisotropy is cross-validated, and with sparse polynomial least squares. In short, the
anisotropy of the method is the dominant factor: with cross-validated anisotropy, both the
Szegö kernel and polynomial least squares attain the pipeline floor, and with the same
anisotropy profile, the Szegö kernel remains about five times more accurate than the
anisotropic Gaussian and Matérn kernels at $N=900$.

\paragraph{Runtime}
On a standard CPU, one evaluation of the trained Szegö surrogate at $N=900$
takes a few milliseconds, whereas a finite-element solve of the same accuracy
takes several tens of milliseconds, so the surrogate is more than an order of
magnitude faster per query. The offline cost of the $900$ training solves is
amortized after a few thousand queries, far fewer than typical Monte Carlo
sample sizes; the gain grows with the cost of the solver, which is low for this
two-dimensional linear problem.

\FloatBarrier
\appendix

\section{Auxiliary estimates}
\label{app:auxiliary_estimates}

This appendix collects the covariance comparison, concentration, and
spectral-factor estimates used in Section~\ref{sec:proof_details}. We use the
notation $\Hs=\Hs_k$, $\I=\I_{\Hs_k}$, and
\[
\Sigma_\lambda=\Sigma+\lambda\I,
\qquad
\hat{\Sigma}_\lambda=\hat{\Sigma}+\lambda\I.
\]

\begin{lemma}\label{lem:comp_betw_lambdas}
If $0<\lambda\leq\lambda_N$, then
\begin{equation}
\label{eq:comp_betw_lambdas}
\norm{\hat{\Sigma}_{\lambda_N}^{1/2}
\hat{\Sigma}_{\lambda}^{-1/2}}_{\Hs\to\Hs}^{2}
\leq
\frac{\lambda_N}{\lambda}.
\end{equation}
\end{lemma}

\begin{proof}
Since $\hat{\Sigma}_{\lambda_N}$ and
$\hat{\Sigma}_{\lambda}$ are functions of the same positive operator
$\hat{\Sigma}$, the spectral theorem gives
\begin{align*}
\norm{\hat{\Sigma}_{\lambda_N}^{1/2}
\hat{\Sigma}_{\lambda}^{-1/2}}_{\Hs\to\Hs}^{2}
&=
\sup_{\sigma\in\operatorname{spec}(\hat{\Sigma})}
\frac{\sigma+\lambda_N}{\sigma+\lambda}\\
&\leq
\frac{\lambda_N}{\lambda},
\end{align*}
because $\lambda_N\geq\lambda$ and
$\sigma\mapsto(\sigma+\lambda_N)/(\sigma+\lambda)$ is decreasing on
$[0,\infty)$.
\end{proof}

The next estimate is a Bernstein type concentration bound for the empirical
and population covariance operators; see, for example,
\cite[Lemma~12]{zhang2024optimality}.

\begin{lemma}\label{lem:bernstein_conc}
Fix $0<\eta<1$ and assume $\Sigma\neq0$. Let $N$ and $\lambda_N$ satisfy
\begin{align}
\label{eq:conc_cond}
u(N,\lambda_N,\eta)
:=
\frac{\mathcal{N}_{\infty}(\lambda_N)}{N}
\log\left(
\frac{
4\mathcal{N}_{\infty}(\lambda_N)
\bigl(\lambda_N+\norm{\Sigma}_{\Hs\to\Hs}\bigr)
}{
\eta\norm{\Sigma}_{\Hs\to\Hs}
}
\right)
\leq \frac{1}{8}.
\end{align}
Then, with probability at least $1-\eta$,
\begin{equation}
\label{eq:resolvent_concentration}
\norm{\Sigma_{\lambda_N}^{1/2}
\hat{\Sigma}_{\lambda_N}^{-1/2}}_{\Hs\to\Hs}^{2}
\leq 3.
\end{equation}
\end{lemma}

Finally, we recall the concentration estimate controlling the empirical
noise term. The displayed bound is a direct consequence of the Bernstein
inequality for Hilbert-space-valued random variables \cite{pinelis1994optimum},
in the form of \cite[Proposition~2]{caponnetto2007optimal}.

\begin{lemma}[Empirical noise concentration]
\label{lem:mom_bound}
Under Assumption~\ref{as:noise}, for every $\lambda>0$ and
$0<\eta<1$, with probability at least $1-\eta$,
\begin{equation}
\label{eq:noise_bound}
\norm{\Sigma_\lambda^{-1/2}
(\hat{\Sigma}C_*-G_z)}_{S_2(\Y,\Hs_k)}
\leq
\frac{4L\log(2/\eta)}{\sqrt{N}}
\left(
\frac{\sqrt{\mathcal{N}_{\infty}(\lambda)}}{\sqrt{N}}
+\sqrt{\mathcal{N}(\lambda)}
\right).
\end{equation}
\end{lemma}

\begin{proof}
Set
\[
Z_i=\Sigma_\lambda^{-1/2}\varphi(x_i)\otimes
\bigl(F_*(x_i)-y_i\bigr).
\]
By Assumption~\ref{as:noise} and the independence of the inputs and the
noise, the variables $Z_i$ are independent and centered. Moreover,
\[
\norm{Z_i}_{S_2(\Y,\Hs_k)}
\leq L\sqrt{\mathcal{N}_{\infty}(\lambda)},
\qquad
\EXP\norm{Z_i}_{S_2(\Y,\Hs_k)}^2
\leq L^2\mathcal{N}(\lambda).
\]
Applying \cite[Proposition~2]{caponnetto2007optimal} with
$M=L\sqrt{\mathcal{N}_{\infty}(\lambda)}$ and $\sigma^2=L^2\mathcal{N}(\lambda)$
to $N^{-1}\sum_{i=1}^N Z_i$ gives the stated bound.
\end{proof}

\begin{lemma}[Logarithmic spectral factor]
\label{lem:log_weight}
Under Assumption~\ref{as:szeg_norm}, for every $t\geq0$ there exists a
constant $C_{t,\kappa}>0$ such that, for every $0<\lambda\leq e^{-1}$,
\begin{equation}
\norm{\tpsi(\Sigma)\Sigma_\lambda^{-1/2}}_{\Hs_k\to\Hs_k}
\leq
C_{t,\kappa}\abs{\log(\lambda)}^t.
\label{eq:log_weight_bound}
\end{equation}
\end{lemma}

\begin{proof}
By the spectral theorem, it is enough to bound
\[
\sup_{0\leq\sigma\leq\kappa^2}
\frac{\sqrt{\sigma}\bigl(1+\abs{\log(\sigma)}\bigr)^t}
{\sqrt{\sigma+\lambda}},
\]
with the convention from Assumption~\ref{as:szeg_norm} at $\sigma=0$.
Since $\lambda\leq e^{-1}$, we have $1+|\log(\lambda)|\leq2|\log(\lambda)|$.
For $0<\sigma\leq\lambda$, the substitution $\sigma=\lambda u$ and the
elementary bound on $u^{1/2}(1+|\log(\lambda u)|)^t$ give a bound by
$C_t|\log(\lambda)|^t$. For $\lambda<\sigma\leq1$, the ratio is bounded
by $(1+|\log(\lambda)|)^t\leq2^t|\log(\lambda)|^t$. On the remaining compact interval, if it is
nonempty, the ratio is bounded by a constant depending only on $t$ and
$\kappa$. This proves~\eqref{eq:log_weight_bound}.
\end{proof}

\begin{lemma}[Power spectral factor]
\label{lem:power_weight}
Under Assumption~\ref{as:sob_norm}, for every $\lambda>0$,
\begin{equation}
\norm{\tpsi(\Sigma)\Sigma_\lambda^{-1/2}}_{\Hs_k\to\Hs_k}
\leq
\lambda^{t-1/2}.
\label{eq:power_weight_bound}
\end{equation}
\end{lemma}

\begin{proof}
For $t=0$, the bound follows from
$\|\Sigma_\lambda^{-1/2}\|\leq\lambda^{-1/2}$. For
$0<t\leq1/2$, the spectral theorem reduces the claim to
\[
\sup_{\sigma\geq0}\frac{\sigma^t}{\sqrt{\sigma+\lambda}}
\leq\lambda^{t-1/2}.
\]
Writing $\sigma=\lambda u$, it is enough to use
$u^t/(1+u)^{1/2}\leq1$ for $0\leq t\leq1/2$ and $u\geq0$.
\end{proof}

\section{Power-series kernels and effective-dimension estimates}
\label{app:analytic_kernels}

This appendix contains the approximation arguments used to estimate the
maximal effective dimension of the Szegö kernel, together with the proofs of
the auxiliary results used in Section~\ref{subsec:szeg}. The power-series
kernels and their associated RKHSs were introduced in
Section~\ref{sec:vrkhs_main}. Throughout, $\X=[-1,1]^P$, and $\mu$ satisfies
Assumption~\ref{as:sampling}.

For a scalar kernel $k$ and its associated RKHS $\Hs_k$, define the integral
operator $T\colon L^2(\X,\mu)\to\Hs_k$ by
\[
(Tw)(\cdot)
=
\int_{\X} k(\cdot,z)w(z)\,\mathrm{d}\mu(z),
\qquad
w\in L^2(\X,\mu).
\]
Let $\varphi(x)=k(x,\cdot)\in\Hs_k$. Then the covariance operator introduced
in Section~\ref{subsec:operator_quantities} satisfies
$\Sigma=TT^*$. For $\lambda>0$, we write
$\Sigma_\lambda=\Sigma+\lambda\I_{\Hs_k}$.

The maximal effective dimension admits the following max--min
representation~\cite[Lemma~5.12]{dommel2024bound}:
\begin{equation}
\mathcal{N}_{\infty}(\lambda)
=
\sup_{x\in\X}
\inf_{w\in L^2(\X,\mu)}
\left\{
\frac{1}{\lambda}
\norm{Tw-\varphi(x)}_{\Hs_k}^2
+
\norm{w}_{L^2(\mu)}^2
\right\}.
\label{eq:max_dof_maxmin}
\end{equation}
Indeed, for fixed $x$, the minimum is attained at
$w=T^*\Sigma_\lambda^{-1}\varphi(x)$, and its value is
$\innerpro{\Sigma_\lambda^{-1}\varphi(x),\varphi(x)}_{\Hs_k}$.

\subsection{Approximation of the feature map}

For the remainder of this appendix, let
$\Hs_k=\Hs_{k_{\mathrm{pow}}}$ and
$\varphi(x)=k_{\mathrm{pow}}(x,\cdot)$. We construct functions with prescribed
polynomial moments.

Let $\mu_1(z)=1/2$ denote the uniform density on $[-1,1]$, and let $P_l$ be
the $l$th Legendre polynomial. Define
$Q_l(z)=\sqrt{2l+1}\,P_l(z)$. Then $(Q_l)_{l\geq0}$ is an orthonormal basis of
$L^2([-1,1],\mu_1)$ and
$\sup_{z\in[-1,1]}|Q_l(z)|\leq\sqrt{2l+1}$.

For $x\in[-1,1]$ and $m\geq1$, define
\[
w_m^x(z)
=
\sum_{l=0}^{m-1}Q_l(x)Q_l(z).
\]
Then, for $k=0,\ldots,m-1$,
\begin{equation}
\int_{-1}^1w_m^x(z)z^k\mu_1(z)\,\mathrm{d}z
=
x^k,
\label{eq:moment}
\end{equation}
and
\begin{equation}
\sup_{x\in[-1,1]}
\norm{w_m^x}_{L^2(\mu_1)}^2
\leq m^2.
\label{eq:uniform}
\end{equation}
These properties follow from the reproducing kernel of the space of
polynomials of degree at most $m-1$; see, for example,
\cite{dommel2024bound}.

Let $\mu_0(z)=2^{-P}$ denote the uniform density on
$\X=[-1,1]^P$ and $\mu(z)$ as in \eqref{bound_dens}. For $x=(x_1,\ldots,x_P)\in\X$, define
\begin{equation}
W_m^x(z)
=
\frac{\mu_0(z)}{\mu(z)}
\prod_{p=1}^P w_m^{x_p}(z_p).
\label{eq:def_weight_multidim}
\end{equation}

\begin{lemma}[Polynomial moment weights]
\label{lem:polynomial_weights}
For every $x\in\X$ and $m\geq1$, the function $W_m^x$ satisfies
\begin{equation}
\int_{\X}z^\nu W_m^x(z)\mu(z)\,\mathrm{d}z
=
x^\nu,
\qquad
\nu\in\mathbb{N}^P,\quad |\nu|<m,
\label{eq:moment_multidim}
\end{equation}
and
\begin{equation}
\sup_{x\in\X}
\norm{W_m^x}_{L^2(\mu)}^2
\leq
C_{\mu,P}m^{2P},
\label{eq:l2_bound_multidim}
\end{equation}
where $C_{\mu,P}>0$ is independent of $m$ and $x$; here and below, $C_{\mu,P}$ denotes a constant depending only on $c_\mu$ and $P$, which may change from line to line.
\end{lemma}

\begin{proof}
Since $|\nu|<m$ implies $\nu_p<m$ for every $p$, the moment identity follows
from cancellation of $\mu$ and $\mu_0$ and separation of variables:
\[
\begin{aligned}
\int_{\X}z^\nu W_m^x(z)\mu(z)\,\mathrm{d}z
&=
\int_{\X}
z^\nu\mu_0(z)
\prod_{p=1}^P w_m^{x_p}(z_p)\,\mathrm{d}z\\
&=
\prod_{p=1}^P
\int_{-1}^1
w_m^{x_p}(z_p)z_p^{\nu_p}\mu_1(z_p)\,\mathrm{d}z_p
=
x^\nu.
\end{aligned}
\]
Moreover, using the lower bound on $\mu$ and
$\mu_0=\prod_{p=1}^P\mu_1$,
\[
\begin{aligned}
\norm{W_m^x}_{L^2(\mu)}^2
&=
\int_{\X}
\frac{\mu_0(z)^2}{\mu(z)}
\prod_{p=1}^P
\left|w_m^{x_p}(z_p)\right|^2\,\mathrm{d}z\\
&\leq
C_{\mu,P}
\prod_{p=1}^P
\int_{-1}^1
\left|w_m^{x_p}(z_p)\right|^2\mu_1(z_p)\,\mathrm{d}z_p\\
&\leq
C_{\mu,P}m^{2P},
\end{aligned}
\]
where the last inequality follows from~\eqref{eq:uniform}.
\end{proof}

For $m\geq1$, define the power-series tail
\begin{equation}
\mathcal{E}_{\mathrm{pow}}(m)
=
\sup_{z\in\X}
\left|
\sum_{|\nu|\geq m}
\frac{w_\nu}{(\nu!)^2}z^\nu
\right|.
\label{eq:tail_pow}
\end{equation}
For a continuous function $h$ on $\X$, we write
$\norm{h}_{\infty}=\sup_{z\in\X}|h(z)|$.

\begin{lemma}
\label{lem:approx_uniform}
For the power-series kernel $k_{\mathrm{pow}}$,
\begin{equation}
\sup_{x\in\X}
\norm{TW_m^x-\varphi(x)}_{\infty}
\leq
\left(1+C_{\mu,P}m^P\right)
\mathcal{E}_{\mathrm{pow}}(m).
\label{eq:approx_uniform}
\end{equation}
\end{lemma}

\begin{proof}
For $x,x'\in\X$, let $x'\odot z$ denote the coordinatewise product of
$x'$ and $z$. The moment property~\eqref{eq:moment_multidim} gives
\[
\begin{aligned}
(TW_m^x)(x')
&=
\int_{\X}k_{\mathrm{pow}}(x',z)W_m^x(z)\mu(z)\,\mathrm{d}z\\
&=
\sum_{|\nu|<m}
\frac{w_\nu}{(\nu!)^2}(x'\odot x)^\nu
+
\int_{\X}
\sum_{|\nu|\geq m}
\frac{w_\nu}{(\nu!)^2}(x'\odot z)^\nu
W_m^x(z)\mu(z)\,\mathrm{d}z.
\end{aligned}
\]
Subtracting
$\varphi(x)(x')=k_{\mathrm{pow}}(x',x)$ yields
\[
(TW_m^x)(x')-\varphi(x)(x')
=
\int_{\X}R_m(x'\odot z)W_m^x(z)\mu(z)\,\mathrm{d}z
-
R_m(x'\odot x),
\]
where
\[
R_m(v)
=
\sum_{|\nu|\geq m}
\frac{w_\nu}{(\nu!)^2}v^\nu.
\]
Since $x'\odot z,x'\odot x\in\X$, the definition of
$\mathcal{E}_{\mathrm{pow}}(m)$ and the Cauchy--Schwarz inequality give
\[
\left|(TW_m^x)(x')-\varphi(x)(x')\right|
\leq
\mathcal{E}_{\mathrm{pow}}(m)
\left(
1+\norm{W_m^x}_{L^2(\mu)}
\right).
\]
Taking the supremum over $x$ and $x'$ and using
\eqref{eq:l2_bound_multidim} proves the claim.
\end{proof}

The uniform approximation estimate also gives an estimate in the RKHS norm.

\begin{lemma}
\label{lem:approx_RKHS}
For the power-series kernel,
\begin{equation}
\sup_{x\in\X}
\norm{TW_m^x-\varphi(x)}_{\Hs_k}^2
\leq
C_{\mu,P}m^{2P}\mathcal{E}_{\mathrm{pow}}(m).
\label{eq:approx_RKHS}
\end{equation}
\end{lemma}

\begin{proof}
Let $e_m^x=TW_m^x-\varphi(x)$. By the definition of $T$ and the
reproducing property,
\[
\norm{e_m^x}_{\Hs_k}^2
=
\innerpro{W_m^x,e_m^x}_{L^2(\mu)}
-
e_m^x(x).
\]
Therefore,
\[
\begin{aligned}
\norm{e_m^x}_{\Hs_k}^2
&\leq
\left|
\innerpro{W_m^x,e_m^x}_{L^2(\mu)}
-
e_m^x(x)
\right|\\
&\leq
\left(
\norm{W_m^x}_{L^2(\mu)}+1
\right)
\norm{e_m^x}_{\infty}.
\end{aligned}
\]
The result follows from Lemma~\ref{lem:approx_uniform} and
\eqref{eq:l2_bound_multidim}.
\end{proof}

\subsection{The Szegö kernel}

For the Szegö kernel, the power-series weights satisfy
\[
\frac{w_\nu}{(\nu!)^2}=c^\nu.
\]
Consequently, since all coefficients are nonnegative,
\begin{equation}
\mathcal{E}_{\mathrm{Szeg\ddot{o}}}(m)
=
\sum_{|\nu|\geq m}c^\nu.
\label{eq:approx_szeg}
\end{equation}

\begin{lemma}
\label{lem:approx_szeg}
Let
\[
c_{\max}=\max_{1\leq p\leq P}c_p<1.
\]
There exists $m_0=m_0(P,c_{\max})$ such that
\begin{equation}
\mathcal{E}_{\mathrm{Szeg\ddot{o}}}(m)
\leq
c_{\max}^{m/2},
\qquad
m\geq m_0.
\label{eq:main_bound_approx}
\end{equation}
\end{lemma}

\begin{proof}
The number of multi-indices $\nu\in\mathbb{N}^P$ with $|\nu|=l$ is
\[
\binom{P+l-1}{l}\leq C_P(l+1)^{P-1}.
\]
Therefore,
\[
\mathcal{E}_{\mathrm{Szeg\ddot{o}}}(m)
\leq
C_P\sum_{l\geq m}(l+1)^{P-1}c_{\max}^l.
\]
Since $c_{\max}<1$, the right-hand side divided by
$c_{\max}^{m/2}$ converges to zero as $m\to\infty$.
Hence, there is $m_0=m_0(P,c_{\max})$ such that
\[
\mathcal{E}_{\mathrm{Szeg\ddot{o}}}(m)
\leq
c_{\max}^{m/2},
\qquad m\geq m_0.
\]
\end{proof}

\begin{proof}[Proof of Proposition~\ref{prop:max_dof_szeg}]
Using the max--min representation~\eqref{eq:max_dof_maxmin} and choosing
$w=W_m^x$, Lemmas~\ref{lem:approx_uniform} and
\ref{lem:approx_RKHS} together with~\eqref{eq:l2_bound_multidim} give
\[
\mathcal{N}_{\infty}(\lambda)
\leq
C_{\mu,P}
\left[
\frac{m^{2P}}{\lambda}
\mathcal{E}_{\mathrm{Szeg\ddot{o}}}(m)
+
m^{2P}
\right].
\]
Choose
\[
m=
\left\lceil
\frac{2}{-\log(c_{\max})}
\log\left(\frac{1}{\lambda}\right)
\right\rceil.
\]
For sufficiently small $\lambda$, this choice satisfies $m\geq m_0$, and
\[
c_{\max}^{m/2}\leq\lambda.
\]
Therefore,
\[
\mathcal{N}_{\infty}(\lambda)
\leq
C_{\mu,P}m^{2P}
\leq
C_{\mu,c_{\max},P}
\left(\log\frac{1}{\lambda}\right)^{2P},
\]
for sufficiently small $\lambda$. This proves the proposition.
\end{proof}

\section{The elliptic solution operator does not belong to the Gaussian VRKHS}
\label{app:gauss}

Example~\ref{ex:gauss} gives an elliptic solution operator that does not
belong to the Gaussian VRKHS. We show here that the same conclusion
holds for every affine diffusion coefficient satisfying
Assumption~\ref{as:pde}, provided that the source term is nonzero.

\begin{proposition}
\label{prop:ell_gaussian_nonmembership}
Let $P\geq1$ and suppose that Assumption~\ref{as:pde} holds, with
real-valued coefficients $\phi_0,\ldots,\phi_P$ and a real-valued source
$f\in L^2(\Omega)$ independent of $x$. As in
Section~\ref{sec:setting}, let $\Y$ be a real separable Hilbert space
with a continuous injective linear map
\[
J:H_0^1(\Omega)\longrightarrow\Y.
\]
For $x\in\X=[-1,1]^P$, let $u_x\in H_0^1(\Omega)$ be the
unique weak solution of~\eqref{eq:elliptic}, with coefficient given by
\eqref{eq:input_coeff}, and set $F_*^{\mathrm{ell}}(x)=Ju_x$.
For  $\bar s>0$, define
\[
K_{\mathrm{Gauss}}(x,x')
=\exp\!\left(-\bar s^2\|x-x'\|_{\mathbb R^P}^2\right)\I_{\Y}.
\]
Then
\[
F_*^{\mathrm{ell}}\in\Hs_{K_{\mathrm{Gauss}}}
\quad\Longleftrightarrow\quad f=0.
\]
\end{proposition}

Recall that every function in the scalar Gaussian RKHS on $[-1,1]$
extends to an entire function. This follows from
\eqref{eq:gauss_rkhs_1d} and the Cauchy--Schwarz inequality; see also
\cite[Theorem~1]{minh2010some}. This space contains no nonzero
constant functions \cite[Theorem~2]{minh2010some}.
We will use these facts together with the following lemma.

\begin{lemma}
\label{lem:entire_selfadjoint_inverse}
Let $V$ be a real Hilbert space, let $T:V\to V$ be bounded and
self-adjoint, and let $b,v\in V$. For real $t$ in a neighborhood of
zero, define
\[
h(t)=\bigl((I_V+tT)^{-1}b,v\bigr)_V.
\]
If $h$ extends to an entire function $H$, then
$H(z)=(b,v)_V$ for every $z\in\mathbb C$.
\end{lemma}

\begin{proof}
Let $E$ be the spectral resolution of $T$ and define
\[
\nu(B)=\bigl(E(B)b,v\bigr)_V,
\qquad B\subseteq\mathbb R\ \text{Borel}.
\]
This is a finite signed measure supported in $[-\|T\|,\|T\|]$.
By the spectral theorem \cite[Section~3.1]{teschl2009mathematical} (applied to the complexification of $V$; $\nu$ is real for real $b,v$),
\[
h(t)=\int_{\mathbb R}\frac{d\nu(\lambda)}{1+t\lambda}
\]
for real $t$ near zero. The integral also defines a holomorphic
function of $t$ in a complex neighborhood of zero, where it agrees
with $H$ by the identity theorem.

Consider the Cauchy transform
\[
C_\nu(z)=\int_{\mathbb R}\frac{d\nu(\lambda)}{z-\lambda},
\qquad z\in\mathbb C\setminus\mathbb R.
\]
Substituting $t=-1/z$ gives
\begin{equation}\label{eq:gaussian_appendix_cauchy}
C_\nu(z)=\frac{H(-1/z)}{z},
\qquad z\in\mathbb C\setminus\mathbb R.
\end{equation}
The identity first holds for large $|z|$ and then throughout each
half-plane by the identity theorem. Since $H$ is real-valued near
zero on the real axis, the identity theorem applied to $H(z)$ and
$\overline{H(\overline z)}$ shows that $H$ is real-valued on
$\mathbb R$. Thus the right-hand side
of~\eqref{eq:gaussian_appendix_cauchy} extends $C_\nu$ holomorphically
to $\mathbb C\setminus\{0\}$ and is real-valued on
$\mathbb R\setminus\{0\}$.

Let $\alpha<\beta$ satisfy $0\notin[\alpha,\beta]$ and
$\nu(\{\alpha\})=\nu(\{\beta\})=0$.
By the Stieltjes inversion formula
\cite[Theorem~3.21]{teschl2009mathematical}, applied by linearity to
the Jordan decomposition of $\nu$,
\[
\nu((\alpha,\beta))
=-\frac{1}{\pi}\lim_{\varepsilon\downarrow0}
\int_\alpha^\beta\operatorname{Im}C_\nu(x+i\varepsilon)\,dx
=0.
\]
The last equality follows because
$\operatorname{Im}C_\nu(x+i\varepsilon)\to0$ uniformly for
$x\in[\alpha,\beta]$. Since $\nu$ has at most countably many atoms,
such endpoints can be chosen densely. It follows that $\nu$ is
supported in $\{0\}$. Therefore
$h(t)=\nu(\mathbb R)=(b,v)_V$ near zero, and the identity theorem
gives the conclusion.
\end{proof}

\begin{namedproof}{Proof of Proposition~\ref{prop:ell_gaussian_nonmembership}}
If $f=0$, uniqueness gives $F_*^{\mathrm{ell}}=0$. We therefore
assume that $f\neq0$.

Equip $V=H_0^1(\Omega)$ with the energy inner product
\[
(u,v)_0=\int_\Omega\phi_0(s)\nabla u(s)\cdot\nabla v(s)\,ds.
\]
By uniform ellipticity, the associated norm is equivalent to the
usual norm on $H_0^1(\Omega)$, so $J:V\to\Y$ is bounded for this
norm. The solution $u_0$ at $x=0$ satisfies
\[
(u_0,v)_0=\int_\Omega f(s)v(s)\,ds,
\qquad v\in V.
\]
Since $H_0^1(\Omega)$ is dense in $L^2(\Omega)$ and $f\neq0$, we
have $u_0\neq0$. Hence $y_0:=Ju_0\neq0$ by injectivity of $J$.

Let $e_1=(1,0,\ldots,0)\in\mathbb R^P$ and restrict the parameters
to $x=te_1$, $t\in[-1,1]$. By the Riesz representation theorem,
there is a bounded operator $T:V\to V$ defined by
\[
(Tu,v)_0=\int_\Omega\phi_1(s)\nabla u(s)\cdot\nabla v(s)\,ds,
\qquad u,v\in V.
\]
The right-hand side is a bounded symmetric bilinear form, so $T$ is
self-adjoint with respect to $(\cdot,\cdot)_0$. For $t\in[-1,1]$,
\[
\bigl((I_V+tT)v,v\bigr)_0
=\int_\Omega a(te_1,s)|\nabla v(s)|^2\,ds
\geq\frac{a_{\mathrm{min}}}{a_{\mathrm{max}}}\|v\|_0^2.
\]
Lax--Milgram therefore implies that $I_V+tT$ is invertible.
The weak formulation of the PDE yields
\begin{equation}\label{eq:gaussian_appendix_slice}
F_*^{\mathrm{ell}}(te_1)=J(I_V+tT)^{-1}u_0,
\qquad t\in[-1,1].
\end{equation}

Suppose, for a contradiction, that
$F_*^{\mathrm{ell}}\in\Hs_{K_{\mathrm{Gauss}}}$.
Restricting the kernel to the line $x=te_1$ and using
\eqref{eq:sep_isom}, we find that
\[
q(t):=\bigl\langle F_*^{\mathrm{ell}}(te_1),y_0\bigr\rangle_\Y
\]
belongs to the scalar Gaussian RKHS on $[-1,1]$ with kernel
$k(t,t')=e^{-\bar s^2(t-t')^2}$. Here we used the standard restriction
property of RKHSs and the fact that, for a separable kernel
$K=k\I_\Y$, taking the inner product with a fixed vector in $\Y$
maps $\Hs_K$ into $\Hs_k$.

Let $J^*:\Y\to V$ denote the adjoint of $J$ for the energy inner
product on $V$, and put $v_0=J^*y_0$. By
\eqref{eq:gaussian_appendix_slice},
\[
q(t)=\bigl((I_V+tT)^{-1}u_0,v_0\bigr)_0.
\]
Since $q$ belongs to the Gaussian RKHS, it extends to an entire
function. By Lemma~\ref{lem:entire_selfadjoint_inverse}, this function
must be constant, with value
\[
q(0)=\langle Ju_0,y_0\rangle_\Y=\|y_0\|_\Y^2>0.
\]
This is impossible, since a nonzero constant does not belong to the
Gaussian RKHS \cite[Theorem~2]{minh2010some}.
\end{namedproof}

\begin{remark}
\label{rem:ell_gaussian_nonmembership_scope}
Assumption~\ref{as:pde_szeg} is not needed for this result. The proof
also does not require the functional $v\mapsto\int_\Omega fv$ on
$H_0^1(\Omega)$ to extend continuously to $\Y$, and it applies even
when the solution is independent of the parameter.

Under~\eqref{bound_dens}, the conclusion also holds when functions
are identified up to equality $\mu$-almost everywhere. Indeed,
$\mu$ has full support on $\X$. Both
$F_*^{\mathrm{ell}}$ and the functions in $\Hs_{K_{\mathrm{Gauss}}}$
are continuous, so equality $\mu$-almost everywhere implies equality
on all of $\X$.
\end{remark}

\section{Further numerical comparisons}
\label{app:further_experiments}

This appendix complements Section~\ref{sec:experimental_overview} with additional baselines
for the parametric elliptic problem with $P=10$. All runs use the data and protocol of
Section~\ref{sec:experimental_overview}: the same ten training pools for each $N$, the same
$1000$ test parameters, the reference mesh $h=1/320$ with bilinear transfer, the relative
$L^2(\Omega)$ error, and five-fold cross-validation with the regularization parameters
$\{10^{-3},10^{-5},10^{-7},10^{-9}\}$. We report the mean relative test error divided by the
pipeline floor; a ratio close to one means that the learning error is negligible compared
with the discretization error of the training data.

\subsection{Anisotropic Gaussian and Matérn kernels}

In Section~\ref{sec:experimental_overview}, the Szegö kernel is anisotropic through
$c_p\propto p^{-2/5}$, whereas the Gaussian and Matérn kernels are isotropic. To separate the
effect of anisotropy from that of the kernel family, we consider coordinate-weighted
(automatic relevance determination, ARD) versions of the Gaussian and Matérn~$\tfrac52$
kernels, based on the distance
$d_w(x,x')^2=\sum_{p=1}^P p^{-2/5}(x_p-x_p')^2$, that is, with the same
coefficient-informed profile as the Szegö family, and cross-validate the overall scale
$\bar s$ on the grid~\eqref{eq:shared_grid}. Conversely, we include an isotropic Szegö
kernel with $c_p=c$ for all $p$, with $c$ cross-validated. The results are given in
Table~\ref{tab:ard}.

\begin{table}[htbp]
\centering
\begin{tabular}{c|cc|cc|cc}
\toprule
 & \multicolumn{2}{c|}{Szegö} & \multicolumn{2}{c|}{Matérn $\tfrac52$} & \multicolumn{2}{c}{Gaussian}\\
$N$ & anisotropic & isotropic & ARD & isotropic & ARD & isotropic\\
\midrule
100 & 1.01 & 1.06 & 1.03 & 1.06 & 1.06 & 1.08\\
300 & 1.04 & 1.45 & 1.37 & 2.22 & 1.49 & 2.38\\
500 & 1.13 & 2.19 & 2.27 & 4.17 & 2.67 & 3.34\\
700 & 1.32 & 3.69 & 4.36 & 7.87 & 5.19 & 6.37\\
900 & 1.63 & 5.98 & 7.83 & 13.85 & 8.17 & 12.50\\
\bottomrule
\end{tabular}
\caption{Mean relative test error divided by the pipeline floor for $P=10$. The anisotropic
Szegö kernel and the isotropic Matérn and Gaussian kernels are those of
Table~\ref{tab:results}; the ARD kernels use the profile $p^{-2/5}$ of the Szegö family.}
\label{tab:ard}
\end{table}

Anisotropy improves every kernel family: the ARD versions reduce the error of the Gaussian
and Matérn kernels by a factor of about $1.5$ to $1.8$ for $N\geq300$. The kernel family
matters in addition. With the same profile, the Szegö kernel is about five times more
accurate than both ARD kernels at $N=900$, and even the isotropic Szegö kernel outperforms
them for $N\geq700$ (paired $t$-tests, $p\leq7\cdot10^{-6}$, better in all ten
repetitions). We used a single fixed profile for the ARD kernels; a coordinate-wise
cross-validation of the ARD weights was not carried out.

\subsection{Cross-validated anisotropy and sparse polynomial least squares}

Sparse polynomial approximation is the classical alternative for parametric elliptic
problems \cite{cohen2010convergence,cohen2011analytic}; see \cite{westermann2026performance}
for a recent comparison of neural and polynomial operator surrogates. As a baseline, we use discrete least
squares in the tensor Legendre basis, orthonormal with respect to the uniform measure on
$[-1,1]^P$, on the downward-closed index set of the $n$ multi-indices with smallest
weighted degree $\sum_{p=1}^P\nu_p(1+\alpha\log p)$; the choice $\alpha=0$ gives the
isotropic total-degree set. The set size $n\in\{5,\dots,650\}$ with $n\leq0.8N$, the
anisotropy $\alpha\in\{0,0.5,1,1.8,3\}$, and a ridge parameter in
$\{0,10^{-10},10^{-8},10^{-6}\}$ are selected by five-fold cross-validation. For a fair
comparison, we also cross-validate the anisotropy of the Szegö kernel, using
$c_p=c_{\mathrm{scale}}\,p^{-2\alpha}$ with $\alpha\in\{0,0.2,0.5,1,1.8,3\}$ and
$c_{\mathrm{scale}}$ from~\eqref{eq:shared_grid}; up to the factor $0.999^2$, $\alpha=0.2$
is the family of Section~\ref{sec:experimental_overview}. The results are given in
Table~\ref{tab:poly}.

\begin{table}[htbp]
\centering
\begin{tabular}{c|cc|cc}
\toprule
 & \multicolumn{2}{c|}{anisotropy cross-validated} & Szegö, $c_p\propto p^{-2/5}$ & polynomial LS,\\
$N$ & Szegö & polynomial LS & (Table~\ref{tab:results}) & isotropic\\
\midrule
100 & 1.000 & 1.000 & 1.013 & 1.12\\
300 & 1.001 & 1.001 & 1.045 & 2.49\\
500 & 1.002 & 1.002 & 1.130 & 3.14\\
700 & 1.003 & 1.007 & 1.325 & 6.39\\
900 & 1.005 & 1.008 & 1.626 & 12.9\\
\bottomrule
\end{tabular}
\caption{Mean relative test error divided by the pipeline floor for $P=10$: Szegö kernel and
sparse polynomial least squares (LS) with cross-validated anisotropy, compared with the
Szegö kernel of Section~\ref{sec:experimental_overview} and isotropic polynomial least
squares.}
\label{tab:poly}
\end{table}

With cross-validated anisotropy, both the Szegö kernel and polynomial least squares attain
the pipeline floor to within one percent for every $N$; the remaining differences are
statistically detectable for $N\geq700$ but of no practical relevance. Cross-validation
selects $\alpha=1.8$ for the polynomial method in all runs and $\alpha\in\{1,1.8\}$ for the
kernel. Without anisotropy, the Szegö kernel is clearly more accurate than isotropic
polynomial least squares, by a factor of about two at $N=900$ ($6.0$ versus $12.9$ times
the floor). On this benchmark, the choice of anisotropy is therefore the dominant factor,
and the Szegö kernel with the coefficient-informed family of
Section~\ref{sec:experimental_overview}, for which the theory applies, remains within a
factor of two of the floor without tuning the anisotropy. The scripts and results for this
appendix are available from the authors upon request.

\bibliographystyle{siam}
\bibliography{bibliography}
\end{document}